\documentclass[10pt,journal,compsoc]{IEEEtran}
\usepackage{color,xcolor}
\usepackage{epsfig}
\usepackage{graphicx}

\usepackage{adjustbox}
\usepackage{array}
\usepackage{booktabs}
\usepackage{colortbl}
\usepackage{float,wrapfig}
\usepackage{hhline}
\usepackage{multirow}
\usepackage{subcaption} %

\usepackage{amsmath,amsfonts,amsthm,amssymb}
\usepackage{bm}
\usepackage{nicefrac}
\usepackage{microtype}

\usepackage{changepage}
\usepackage{extramarks}
\usepackage{fancyhdr}
\usepackage{lastpage}
\usepackage{setspace}
\usepackage{soul}
\usepackage{xspace}
\usepackage{pifont}
\usepackage{cuted}
\usepackage{wrapfig}

\usepackage[pagebackref=true,breaklinks=true,letterpaper=true,colorlinks,  bookmarks=false]{hyperref}
\usepackage{url}

\usepackage{algorithm}
\usepackage{algpseudocode}
\usepackage{enumitem}

\usepackage{wasysym}
\newcolumntype{L}[1]{>{\raggedright\let\newline\\\arraybackslash\hspace{0pt}}m{#1}}
\newcolumntype{C}[1]{>{\centering\let\newline\\\arraybackslash\hspace{0pt}}m{#1}}
\newcolumntype{R}[1]{>{\raggedleft\let\newline\\\arraybackslash\hspace{0pt}}m{#1}}

\newcommand{\sect}[1]{Section~\ref{#1}}

\newcommand{\eqn}[1]{Equation~\ref{#1}}

\newcommand{\tab}[1]{Table~\ref{#1}}

\newcommand{\x}{\mathbf{x}}
\newcommand{\y}{\mathbf{y}}
\newcommand{\fid}{Fr\'echet Inception Distance\xspace}

\newcommand{\lblsec}[1]{\label{sec:#1}}

\newcommand{\ignorethis}[1]{}
\newcommand{\norm}[1]{\lVert#1\rVert_1}

\DeclareMathOperator*{\argmin}{arg\,min}

\makeatletter
\DeclareRobustCommand\onedot{\futurelet\@let@token\@onedot}
\def\@onedot{\ifx\@let@token.\else.\null\fi\xspace}

\def\eg{\emph{e.g}\onedot} 
\def\ie{\emph{i.e}\onedot} 
 
 \def\vs{\emph{vs}\onedot}
 
\def\etal{\emph{et al}\onedot}
\makeatother

\definecolor{citecolor}{RGB}{34,139,34}
\definecolor{mydarkblue}{rgb}{0,0.08,1}
\definecolor{mydarkgreen}{rgb}{0.02,0.6,0.02}
\definecolor{mydarkred}{rgb}{0.8,0.02,0.02}
\definecolor{mydarkorange}{rgb}{0.40,0.2,0.02}
\definecolor{mypurple}{RGB}{111,0,255}
\definecolor{myred}{rgb}{1.0,0.0,0.0}
\definecolor{mygold}{rgb}{0.75,0.6,0.12}
\definecolor{mydarkgray}{rgb}{0.66, 0.66, 0.66}

\def\ndatasets{five\xspace}
\def\multirowcenter{-0.5\dimexpr \aboverulesep + \belowrulesep + \cmidrulewidth}

\ifCLASSOPTIONcompsoc
  \usepackage[nocompress]{cite}
\else
  \usepackage{cite}
\fi

\ifCLASSINFOpdf
\else
\fi
\begin{document}
\title{GAN Compression: Efficient Architectures for Interactive Conditional GANs}
\author{Muyang Li, Ji Lin, Yaoyao Ding, Zhijian Liu, Jun-Yan Zhu and Song Han\IEEEcompsocitemizethanks{\IEEEcompsocthanksitem M. Li and J.-Y. Zhu are with Carnegie Mellon University. \protect E-mail: \{muyangli,junyanz\}@cs.cmu.edu\IEEEcompsocthanksitem J. Lin, Z. Liu and S. Han are with the Department of Electrical Engineering and Computer Science, Massachusetts Institute of Technology. \protect E-mail: \{jilin, zhijian, songhan\}@mit.edu\IEEEcompsocthanksitem Y. Ding is with University of Toronto. \protect E-mail: yaoyao.ding@mail.utoronto .ca}
}

% The paper headers
\markboth{IEEE TRANSACTIONS ON PATTERN ANALYSIS AND MACHINE INTELLIGENCE, VOL. X, NO. X, MMMMMMM YYYY}{Li \MakeLowercase{\textit{et al.}}: GAN Compression: Efficient Architectures for Interactive Conditional GANs}
\IEEEtitleabstractindextext{
\begin{abstract}
Conditional Generative Adversarial Networks (cGANs) have enabled controllable image synthesis for many vision and graphics applications. 
However, recent cGANs are 1-2 orders of magnitude more compute-intensive than modern recognition CNNs. 
For example, GauGAN consumes 281G MACs per image, compared to 0.44G MACs for MobileNet-v3, making it difficult for interactive deployment. 
In this work, we propose a general-purpose compression framework for reducing the inference time and model size of the generator in cGANs.
Directly applying existing compression methods yields poor performance due to the difficulty of GAN training and the differences in generator architectures. We address these challenges in two ways. 
First, to stabilize GAN training, we transfer knowledge of multiple intermediate representations of the original model to its compressed model and unify unpaired and paired learning. 
Second, instead of reusing existing CNN designs, our method finds efficient architectures via neural architecture search. To accelerate the search process, we decouple the model training and search via weight sharing.
Experiments demonstrate the effectiveness of our method across different supervision settings, network architectures, and learning methods. 
Without losing image quality, we reduce the computation of CycleGAN by 21$\times$, Pix2pix by 12$\times$, MUNIT by 29$\times$, and GauGAN by 9$\times$, paving the way for interactive image synthesis. 

\end{abstract}

}

\maketitle

\IEEEdisplaynontitleabstractindextext
\IEEEpeerreviewmaketitle

\begin{strip}
\vspace{-95pt}
\centering
\includegraphics[width=\linewidth]{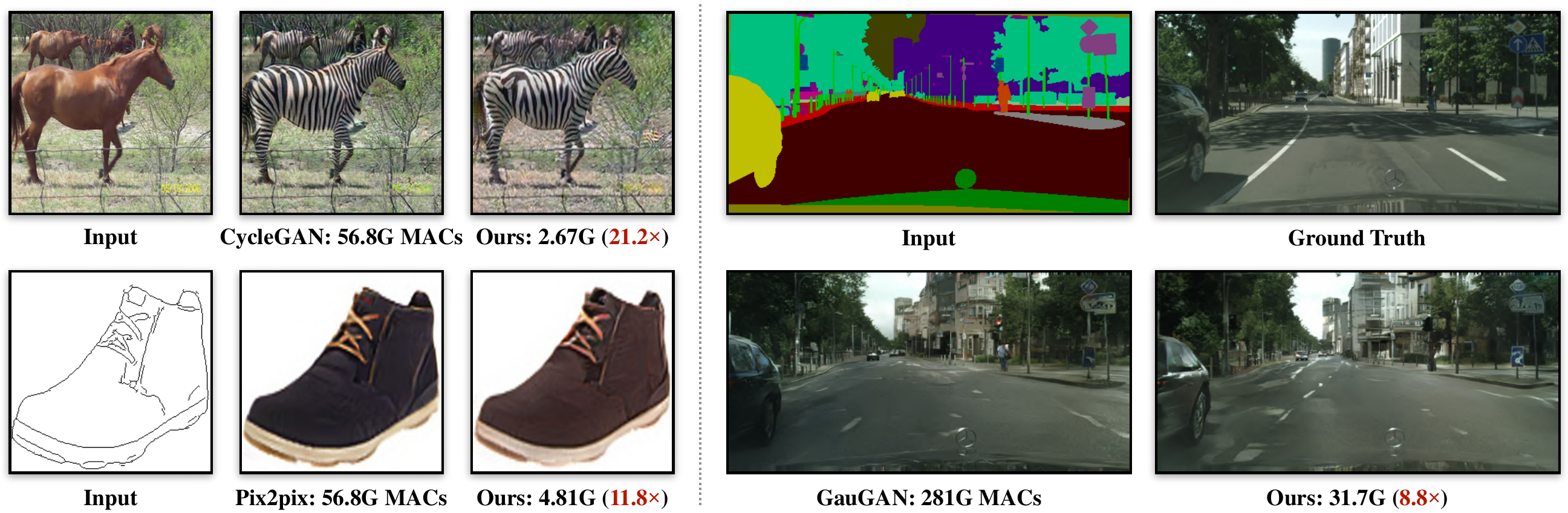}
\captionof{figure}{We introduce \emph{GAN Compression}, a general-purpose method for compressing conditional GANs. Our method reduces the computation of widely-used conditional GAN models including Pix2pix, CycleGAN, and GauGAN by 9-21$\times$ while preserving the visual fidelity. Our method is effective for a wide range of generator architectures, learning objectives, and both paired and unpaired settings.}  
\vspace{5pt}
\label{fig:teaser}
\end{strip}

\IEEEraisesectionheading{\section{Introduction}}

\IEEEPARstart{G}{enerative} Adversarial Networks (GANs)~\cite{goodfellow2014generative} excel at synthesizing photo-realistic images. Their conditional extension, conditional GANs~\cite{mirza2014conditional,isola2017image,zhu2017unpaired}, allows controllable image synthesis and enables many computer vision and graphics applications such as interactively creating an image from a user drawing~\cite{park2019semantic}, transferring the motion of a dancing video stream to a different person~\cite{wang2018video,chan2019everybody,aberman2019learning}, or creating VR facial animation for remote social
interaction~\cite{wei2019vr}. All of these applications require models to interact with humans and therefore demand low-latency on-device performance for a better user experience. However, edge devices (mobile phones, tablets, VR headsets) are tightly constrained by hardware resources such as memory and battery. This computational bottleneck prevents conditional GANs from being deployed on edge devices. 

Different from image recognition CNNs~\cite{krizhevsky2012imagenet,simonyan2015very,he2016deep,howard2017mobilenets}, image-conditional GANs are notoriously computationally intensive. For example, the widely-used CycleGAN model~\cite{zhu2017unpaired} requires more than 50G MACs\footnote{We use the number of Multiply-Accumulate Operations (MAC) to quantify the computation cost. Modern computer architectures use fused multiply-add (FMA) instructions for tensor operations. These instructions compute $a=a+b\times c$ as one operation. 1 MAC=2 FLOPs.}, 100$\times$ more than MobileNet~\cite{howard2017mobilenets,sandler2018mobilenetv2}. A more recent model GauGAN~\cite{park2019semantic}, though generating photo-realistic high-resolution images, requires more than 250G MACs, 500$\times$ more than MobileNet~\cite{howard2017mobilenets,sandler2018mobilenetv2}. 

In this work, we present \emph{GAN Compression}, a general-purpose compression method for reducing the inference time and computational cost for conditional GANs. We observe that compressing generative models faces two fundamental difficulties:  GANs are unstable to train, especially under the unpaired setting; generators also differ from recognition CNNs, making it hard to reuse existing CNN designs. To address these issues, we first transfer the knowledge from the intermediate representations of the original teacher generator to the corresponding layers of its compressed student generator. We also find it beneficial to create pseudo pairs using the teacher model's output for unpaired training. This transforms unpaired learning to paired learning. Second, we use neural architecture search (NAS) to automatically find an efficient network with significantly fewer computation costs and parameters. To reduce the training cost, we decouple the model training from architecture search by training a once-for-all network that contains all possible channel number configurations. The once-for-all network can generate many sub-networks by weight sharing and enable us to evaluate the performance of each sub-network without retraining. Our method can be applied to various conditional GAN models regardless of model architectures, learning algorithms, and supervision settings (paired or unpaired).

Through extensive experiments, we show that our method can reduce the computation of four widely-used conditional GAN models, including Pix2pix~\cite{isola2017image}, CycleGAN~\cite{zhu2017unpaired}, GauGAN~\cite{park2019semantic}, and MUNIT~\cite{huang2018multimodal}, by 9$\times$ to 29$\times$ regarding MACs, without loss of the visual fidelity of generated images (see Figure~\ref{fig:teaser} for several examples). 
Finally, we deploy our compressed pix2pix model on a mobile device (Jetson Nano) and demonstrate an interactive edges$\to$shoes application (\href{https://youtu.be/31AhcLqWc68}{demo}), paving the way for interactive image synthesis.

Compared to our conference version, this paper includes new development and extensions in the following aspects:
\begin{itemize}
	\item We introduce Fast GAN Compression, a more efficient training method with a simplified training pipeline and a faster search strategy. Fast GAN Compression reduces the training time of original GAN Compression by 1.7-3.7$\times$ and the search time by 3.5-12$\times$. 
	\item To demonstrate the generality of our proposed method, we test our methods on additional datasets and models. We first compress GauGAN on the challenging COCO-Stuff dataset, achieving 5.4$\times$ computation reduction. We then apply Fast GAN Compression to MUNIT on the edges$\to$shoes dataset. Although MUNIT uses a different type of architectures (\ie, a style encoder, a content encoder, and a decoder) and training objectives compared to CycleGAN, our method still manages to reduce the computation by 29$\times$ and model size by 15$\times$, without loss of visual fidelity. 
	\item We show additional evaluation based on the perceptual similarity metric, user study, and per-class analysis on Cityscapes. We also include extra ablation studies regarding individual components and design choices. 
\end{itemize}

Our \href{https://github.com/mit-han-lab/gan-compression}{code} and \href{https://www.youtube.com/playlist?list=PL80kAHvQbh-r5R8UmXhQK1ndqRvPNw_ex}{demo} are publicly available.
\section{Related Work}

\subsection{Conditional GANs}
Generative Adversarial Networks (GANs)~\cite{goodfellow2014generative} excel at synthesizing photo-realistic results~\cite{karras2019style,brock2018large}. Its conditional form, conditional GANs~\cite{mirza2014conditional,isola2017image}, further enables controllable image synthesis, allowing a user to synthesize images given various conditional inputs such as user sketches~\cite{isola2017image,sangkloy2017scribbler}, class labels~\cite{mirza2014conditional,brock2018large}, or textual descriptions~\cite{reed2016generative,zhang2018stackgan++}. Subsequent works further increase the resolution and realism of the results~\cite{wang2018high,park2019semantic}. Later, several algorithms were proposed to learn conditional GANs without paired data~\cite{taigman2017unsupervised,shrivastava2017learning,zhu2017unpaired,kim2017learning,yi2017dualgan,liu2017unsupervised,choi2018stargan,huang2018multimodal,lee2018diverse}.

\begin{figure}[t]
\centering
\includegraphics[width=\linewidth]{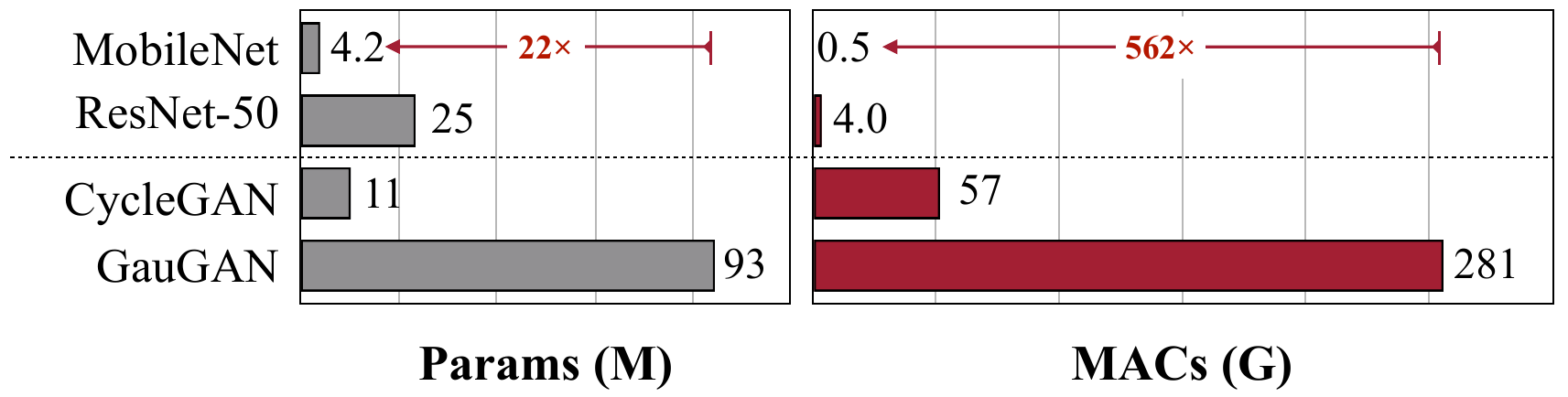}
\caption{Conditional GANs require two orders of magnitude (562$\times$) more computation than image classification CNNs, making it prohibitive to be deployed on edge devices.}  
\vspace{-15pt}
\label{fig:related}
\end{figure}

The high-resolution, photo-realistic synthesized results come at the cost of intensive computation. As shown in Figure~\ref{fig:related}, although the model size is of the same magnitude as the size of image recognition CNNs~\cite{he2016deep}, conditional GANs require two orders of magnitudes more computations. This makes it challenging to deploy these models on edge devices given limited computational resources. In this work, we focus on efficient image-conditional GANs architectures for interactive applications. 

\subsection{Model Acceleration}
Extensive attention has been paid to hardware-efficient deep learning for various real-world applications~\cite{han2015learning,han2015deep,zhu2016trained,wang2019haq,han2019design}. To reduce redundancy in network weights, researchers proposed to prune the connections between layers~\cite{han2015learning,han2015deep,wen2016learning}. However, the pruned networks require specialized hardware to achieve their full speedup. Several subsequent works proposed to prune entire convolution filters~\cite{he2017channel,lin2017runtime,liu2017learning} to improve the regularity of computation. AutoML for Model Compression (AMC)~\cite{he2018amc} leverages reinforcement learning to determine the pruning ratio of each layer automatically. Liu~\etal~\cite{liu2019metapruning} later replaced reinforcement learning with an evolutionary search algorithm. 
Previously, Shu~\etal~\cite{shu2019co} proposed co-evolutionary pruning for CycleGAN by modifying the original CycleGAN algorithm. This method is tailored for a particular algorithm. The compressed model significantly increases FID under a moderate compression ratio (4.2$\times$). In contrast, our model-agnostic method can be applied to conditional GANs with different learning algorithms, architectures, and both paired and unpaired settings. We assume no knowledge of the original cGAN learning algorithm. Experiments show that our general-purpose method achieves a 21.1$\times$ compression ratio (5$\times$ better than the CycleGAN-specific method~\cite{shu2019co}) while retaining the FID of original models. 
Since the publication of our preliminary conference version, several methods on GANs compression and acceleration have been proposed. 
These include different applications such as compressing unconditional GANs~\cite{lin2021anycost,liu2021content,hou2021slimmable} and text-to-image GANs~\cite{ma2021cpgan}, and different algorithms, including efficient generator architectures~\cite{shaham2020spatially}, neural architecture search ~\cite{fu2020autogan,li2020learning}, and channel pruning~\cite{fu2020autogan,li2020learning,jin2021teachers,wang2020gan}.

\subsection{Knowledge Distillation}
\begin{figure*}[h]
\centering
\vspace{-5pt}
\includegraphics[width=1.0\linewidth]{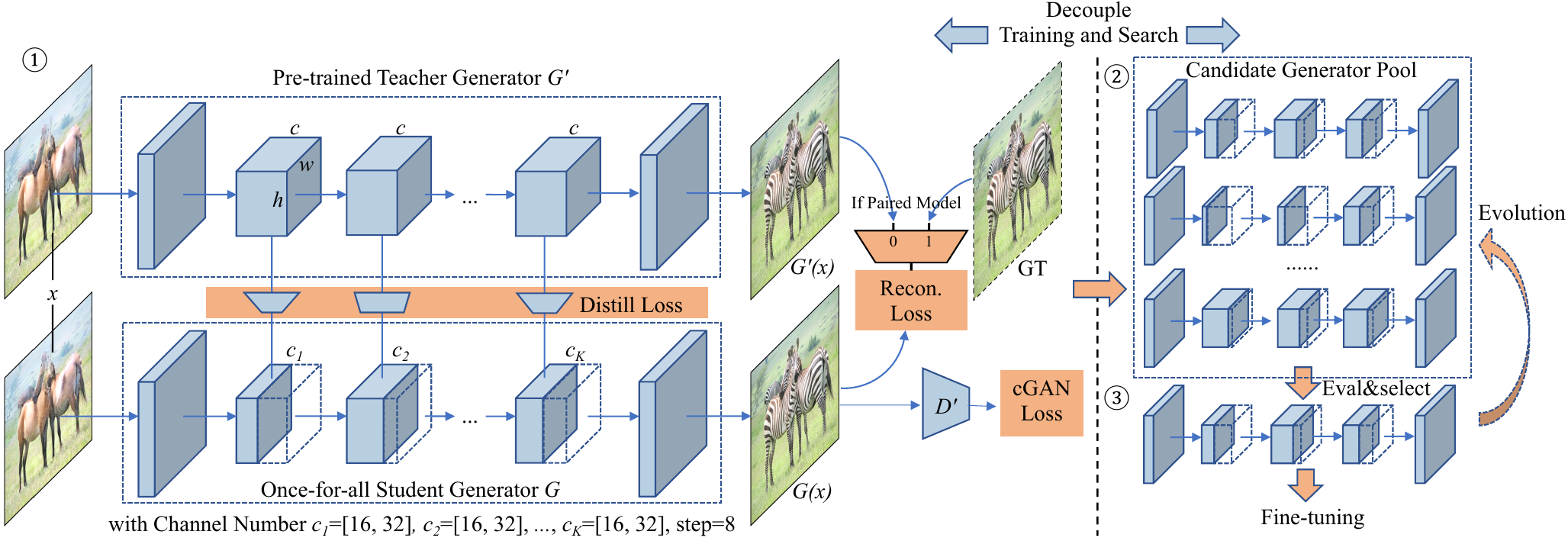}
\caption{GAN Compression framework: {\large \ding{192}} Given a pre-trained teacher generator $G'$, we distill a smaller once-for-all student generator $G$ that contains all possible channel numbers through weight sharing. We choose different channel numbers $\{c_k\}_{k=1}^K$ for the student generator $G$ at each training step. {\large \ding{193}} We then extract many sub-generators from the once-for-all generator and evaluate their performance. No retraining is needed, which is the advantage of the once-for-all generator. {\large \ding{194}} Finally, we choose the best sub-generator given the compression ratio target or performance target (FID or mIoU) using either brute-force or evolutionary search method. Optionally, we perform additional fine-tuning to obtain the final model.
}
\vspace{-10pt}
\label{fig:overall}
\end{figure*}

Hinton~\etal~\cite{hinton2015distilling} introduced the knowledge distillation for transferring the knowledge in a larger teacher network to a smaller student network. The student network is trained to mimic the behavior of the teacher network. Several methods leverage knowledge distillation for compressing recognition models~\cite{luo2016face,chen2017learning,li2018knowledge}. Recently, Aguinaldo~\etal~\cite{aguinaldo2019compressing} adopted this method to accelerate unconditional GANs. Different from them, we focus on conditional GANs. 
We experimented with several distillation methods~\cite{aguinaldo2019compressing,yim2017gift} on conditional GANs and only observed marginal improvement, insufficient for interactive applications. Please refer to Section~\ref{sec:distillation} for more details.

\subsection{Neural Architecture Search}

Neural Architecture Search (NAS) has successfully designed neural network architectures that outperform hand-crafted ones for large-scale image classification tasks~\cite{zoph2016neural,liu2018progressive,liu2018hierarchical}. To effectively reduce the search cost, researchers recently proposed one-shot neural architecture search~\cite{liu2018darts,cai2018proxylessnas,wu2019fbnet,guo2020single,howard2019searching,bender2018understanding, cai2020once} in which different candidate sub-networks can share the same set of weights. While all of these approaches focus on image classification models, we study efficient conditional GANs architectures using NAS.

\section{Method}
Compressing conditional generative models for interactive applications is challenging due to two reasons. Firstly, the training dynamic of GANs is highly unstable by nature. Secondly, the large architectural differences between recognition and generative models make it hard to apply existing CNN compression algorithms directly. To address the above issues, we propose a training protocol tailored for efficient generative models (\sect{sec:method_objective}) and further increase the compression ratio with neural architecture search (NAS) (\sect{sec:method_surgery}). In \sect{sec:pipelines}, we introduce a Fast GAN Compression pipeline, which further reduces the training time of our original compression method. The overall framework is illustrated in Figure~\ref{fig:overall}. Here, we use the ResNet generator~\cite{johnson2016perceptual,zhu2017unpaired} as an example. However, the same framework can be applied to different generator architectures and learning objectives.

\subsection{Training Objective}
\lblsec{method_objective}
\subsubsection{Unifying Unpaired and Paired Learning} 
Conditional GANs aim to learn a mapping function $G$ between a source domain $X$ and a target domain $Y$. They can be trained using either \emph{paired} data ($\{\x_{i}, \y_{i}\}_{i=1}^N$ where $\x_{i} \in X$ and $\y_{i} \in Y$) or \emph{unpaired} data (source dataset $\{\x_{i}\}^N_{i=1}$ to target dataset $\{\y_{j}\}^M_{j=1}$). Here, $N$ and $M$ denote the number of training images.  For simplicity, we omit the subscript $i$ and $j$. Several learning objectives have been proposed to handle both paired and unpaired settings~\cite{isola2017image,park2019semantic,wang2018high,zhu2017unpaired,liu2017unsupervised,huang2018multimodal}.
The wide range of training objectives makes it difficult to build a general-purpose compression framework. To address this, we unify the unpaired and paired learning in the model compression setting, regardless of how the teacher model is originally trained. 
Given the original teacher generator $G'$, we can transform the unpaired training setting to the paired setting.
In particular, for the unpaired setting, we can view the original generator's output as our ground-truth and train our compressed generator $G$ with a paired learning objective. Our learning objective can be summarized as follows:

\begin{equation}
    \label{eqn:loss_recon}
    \mathcal{L}_\text{recon} = \begin{cases}
        \mathbb{E}_{\x, \y} \norm{G(\x) - \y}  & \text{for paired cGANs}, \\
        \mathbb{E}_{\x} \norm{G(\x) - G'(\x)}  & \text{for unpaired cGANs}. \\
    \end{cases}
\end{equation}
Here we denote $\mathbb{E}_{\x} \triangleq \mathbb{E}_{\x\sim p_{\text{data}}(\x)}$ and $ \mathbb{E}_{\x,\y} \triangleq \mathbb{E}_{\x,\y \sim p_{\text{data}}(\x, \y)}$ for simplicity. $\norm{}$ denotes L1 norm. 

With such modifications, we can apply the same compression framework to different types of cGANs.
Furthermore, as shown in Section~\ref{sec:unpaired-to-paired}, learning using the above pseudo pairs makes training more stable and yields much better results compared to the original unpaired training setting.

As the unpaired training has been transformed into paired training, we will discuss the following sections in the paired training setting unless otherwise specified.

\subsubsection{Inheriting the Teacher Discriminator}

Although we aim to compress the generator, the discriminator $D$ stores useful knowledge of a learned GAN as $D$ learns to spot the weakness of the current generator~\cite{azadi2018discriminator}. Therefore, we adopt the same discriminator architecture, use the pre-trained weights from the teacher, and fine-tune the discriminator together with our compressed generator. In our experiments, we observe that a pre-trained discriminator could guide the training of our student generator. Using a randomly initialized discriminator often leads to severe training instability and the degradation of image quality. The GAN objective is formalized as:
\begin{equation}
    \label{eqn:loss_gan}
    \mathcal{L}_\text{cGAN} = \mathbb{E}_{\x, \y}[\log D(\x, \y)] + \mathbb{E}_{\x}[\log (1 - D(\x, G(\x)))],
\end{equation}
where we initialize the student discriminator $D$ using the weights from teacher discriminator $D'$. $G$ and $D$ are trained using a standard minimax optimization~\cite{goodfellow2014generative}.

\subsubsection{Intermediate Feature Distillation}
\label{sec:body-distillation}
A widely-used method for CNN model compression is knowledge distillation~\cite{hinton2015distilling,luo2016face,chen2017learning,yim2017gift,li2018knowledge,polino2018model,chen2018darkrank}. By matching the distribution of the output layer's logits, we can transfer the dark knowledge from a teacher model to a student model, improving the performance of the student. However, conditional GANs~\cite{isola2017image,zhu2017unpaired} usually output a deterministic image rather than a probabilistic distribution. Therefore, it is difficult to distill the dark knowledge from the teacher's output pixels. Especially for paired training setting,  output images generated by the teacher model essentially contains no additional information compared to ground-truth target images. 
Experiments in Section~\ref{sec:distillation} show that for paired training, naively mimicking the teacher model's output brings no improvement.

To address the above issue, we match the intermediate representations of the teacher generator instead, as explored in prior work~\cite{li2018knowledge,zagoruyko2017paying,chen2017learning}. The intermediate layers contain more channels, provide richer information, and allow the student model to acquire more information in addition to  outputs. The distillation objective can be formalized as
\begin{equation}
    \label{eqn:loss_distill}
    \mathcal{L}_\text{distill} = \sum_{t=1}^T{\|{f_t(G_t(\x)) - G'_t(\x)\|_2}},
\end{equation}
where $G_t(\x)$ and $G'_t(\x)$ are the intermediate feature activations of the $t$-th chosen layer in the student and teacher models, and $T$ denotes the number of layers. A 1$\times$1 learnable convolution layer  $f_t$ maps the features from the student model to the same number of channels in the features of the teacher model. We jointly optimize $G_t$ and $f_t$ to minimize the distillation loss $\mathcal{L}_\text{distill}$. Section~\ref{subsubsec:distillation layers} details intermediate distillation layers we choose for each generator architecture in our experiments.

\subsubsection{Full Objective}
Our final objective is written as follows:
\begin{equation}
    \label{eqn:loss}
    \mathcal{L} = \mathcal{L}_\text{cGAN} + \lambda_{\text{recon}} \mathcal{L}_\text{recon} + \lambda_{\text{distill}} \mathcal{L}_\text{distill},
\end{equation}
where hyper-parameters $\lambda_{\text{recon}}$ and $\lambda_{\text{distill}}$ control the importance of each term. Please refer to our \href{https://github.com/mit-han-lab/gan-compression}{code} for more details. 

\subsection{Efficient Generator Design Space}
\lblsec{method_surgery}

Choosing a well-designed student architecture is essential for the final performance of knowledge distillation. We find that naively shrinking the channel numbers of the teacher model fails to produce a compact student model: the performance starts to degrade significantly above 4$\times$ computation reduction.
One of the possible reasons is that existing generator architectures are often adopted from image recognition models~\cite{long2015fully,he2016deep,ronneberger2015u}, and may not be the optimal choice for image synthesis tasks. Below, we show how we derive a better architecture design space from an existing cGAN generator and perform neural architecture search (NAS) within the space. 

\subsubsection{Convolution Decomposition and Layer Sensitivity}
Existing generators usually adopt vanilla convolutions to follow the design of classification and segmentation CNNs. Recent efficient CNN designs widely adopt a decomposed version of convolutions (depthwise + pointwise)~\cite{howard2017mobilenets}, which proves to have a better performance-computation trade-off. We find that using the decomposed convolution also benefits the generator design in cGANs. 

Unfortunately, our early experiments have shown that directly decomposing all the convolution layers (as in classifiers) will significantly degrade the image quality. Decomposing some of the layers will immediately hurt the performance, while other layers are robust. Furthermore, this layer sensitivity pattern is not the same as recognition models. For example, for the ResNet generator~\cite{he2016deep,johnson2016perceptual} in CycleGAN and Pix2pix, the ResBlock layers consume the majority of the model parameters and computation cost while is almost immune to decomposition. On the contrary, the upsampling layers have much fewer parameters but are fairly sensitive to model compression: moderate compression can lead to a large FID degradation. Therefore, we only decompose the ResBlock layers. We conduct a comprehensive study regarding the sensitivity of layers in Section~\ref{sec:decomposition}. See Section~\ref{subsubsec:mobile arch} for more details about the MobileNet-style architectures for other generators in our experiments.

\subsubsection{Automated Channel Reduction with NAS}
Existing generators use hand-crafted (and mostly uniform) channel numbers across all the layers, which contains redundancy and is far from optimal. To further improve the compression ratio, we automatically select the channel width in the generators using channel pruning~\cite{he2017channel,he2018amc,liu2017learning,zhuang2018discrimination,luo2017thinet} to remove the redundancy, which can reduce the computation quadratically. 
 We support fine-grained choices regarding the numbers of channels. For each convolution layer, the number of channels can be chosen from multiples of 8, which balances MACs and hardware parallelism~\cite{he2018amc}.

Given the possible channel configurations $\{c_1, c_2, ..., c_K\}$, where $K$ is the number of layers to prune, our goal is to find the best channel configuration $\{c_1^*, c_2^*, ..., c_K^*\} = \argmin_{c_1, c_2, ..., c_K}\mathcal{L}, \quad s.t.\;\text{MACs}<F$ using neural architecture search, where $F$ is the computation budget. This budget is chosen according to the hardware constraint and target latency. A straightforward method 
for choosing the best configuration is to traverse all the possible channel configurations, train them to convergence, evaluate, and pick the generator with the best performance.
However, as $K$ increases, the number of possible configurations increases exponentially, and each configuration might require different hyper-parameters regarding the learning rates and weights for each term. 
This trial and error process is far too time-consuming.

\subsubsection{Decouple Training and Search}
\label{sec:decouple}
To address the problem, we decouple model training from  architecture search, following recent work in one-shot neural architecture search methods~\cite{cai2018proxylessnas,cai2020once,guo2020single}. 
We first train a once-for-all network~\cite{cai2020once} that supports different channel numbers. Each sub-network with different numbers of channels is equally trained and can operate independently. Sub-networks share the weights within the once-for-all network. 
Figure~\ref{fig:overall} illustrates the overall framework. We assume that the original teacher generator has $\{c_k^0\}_{k=1}^K$ channels. For a given channel number configuration $\{c_k\}_{k=1}^K, c_k\leq c_k^0$, we obtain the weight of the sub-network by extracting the first $\{c_k\}_{k=1}^K$ channels from the corresponding weight tensors of the once-for-all network, following Guo \etal~\cite{guo2020single}. 
At each training step, we randomly sample a sub-network with a certain channel number configuration, compute the output and gradients, and update the extracted weights using our learning objective (\eqn{eqn:loss}). Since the weights at the first several channels are updated more frequently, they play a more critical role among all the weights.

\begin{table*}[t]
\renewcommand*{\arraystretch}{1.15}
\setlength{\tabcolsep}{3.5pt}
\small\centering
\begin{tabular}{cclcccccccc}
\toprule
\multirow{2}{*}[\multirowcenter]{Model} & \multirow{2}{*}[\multirowcenter]{Dataset} & \multirow{2}{*}[\multirowcenter]{Method} & \multicolumn{2}{c}{\#Parameters} & \multicolumn{2}{c}{MACs} & \multicolumn{2}{c}{FID ($\downarrow$)} & \multicolumn{2}{c}{mIoU ($\uparrow$)}\\ 
\cmidrule(lr){4-5} \cmidrule(lr){6-7} \cmidrule(lr){8-9} \cmidrule(lr){10-11}
& & & Value & Ratio & Value & Ratio & Value & Increase & Value & Drop \\
\midrule
\multirow{5}{*}{CycleGAN}& \multirow{5}{*}{horse$\to$zebra}& Original & 11.4M & -- & 56.8G & -- & 61.53 & -- & \multicolumn{2}{c}{--} \\
& & Shu~\etal~\cite{shu2019co} & -- & -- & 13.4G & 4.2$\times$ & 96.15 & 34.6 & \multicolumn{2}{c}{--} \\
& & Ours (w/o fine-tuning) & \textbf{0.34M} & \textbf{33.3$\times$} & 2.67G & 21.2$\times$ & \textbf{64.95} & \textbf{3.42} & \multicolumn{2}{c}{--} \\
& & Ours & 0.34M & 33.3$\times$ & 2.67G & 21.2$\times$ & 71.81 & 10.3 & \multicolumn{2}{c}{--} \\
& & Ours (Fast) & 0.36M & 32.1$\times$ & \textbf{2.64G} & \textbf{21.5$\times$} & 65.19 & 3.66 & \multicolumn{2}{c}{--} \\
\midrule
\multirow{5}{*}{MUNIT}& \multirow{5}{*}{edges$\to$shoes} & Original & 15.0M & -- & 77.3G & -- & 30.13 & -- & \multicolumn{2}{c}{--} \\
& & Ours (w/o fine-tuning) & \textbf{1.00M} & \textbf{15.0}$\times$ & \textbf{2.58G} & \textbf{30.0}$\times$ & \textbf{32.81} & \textbf{2.68} & \multicolumn{2}{c}{--} \\
& & Ours & 1.00M & 15.0$\times$ & 2.58G & 30.0$\times$ & 31.48 & 1.35 & \multicolumn{2}{c}{--} \\
& & Ours (Fast, w/o fine-tuning) & \textbf{1.10M} & \textbf{13.6}$\times$ & \textbf{2.63G} & \textbf{29.4}$\times$ & \textbf{30.53} & \textbf{0.40} & \multicolumn{2}{c}{--} \\
& & Ours (Fast) & 1.10M & 13.6$\times$ & 2.63G & 29.4$\times$ & 32.51 & 1.98  & \multicolumn{2}{c}{--} \\\midrule
 & \multirow{4}{*}{edges$\to$shoes} & Original & 11.4M & -- & 56.8G & -- & 24.18 & -- & \multicolumn{2}{c}{--} \\
& & Ours (w/o fine-tuning) & 0.70M & 16.3$\times$ & 4.81G & 11.8$\times$ & 31.30 & 7.12 & \multicolumn{2}{c}{--}\\
& & Ours & \textbf{0.70M} & \textbf{16.3$\times$} & \textbf{4.81G} & \textbf{11.8$\times$} & 26.60 & 2.42 & \multicolumn{2}{c}{--} \\
& & Ours (Fast) & 0.70M & 16.3$\times$ & 4.87G & 11.7$\times$ & \textbf{25.76} & \textbf{1.58} & \multicolumn{2}{c}{--} \\
\cmidrule{2-11}
\multirow{4}{*}{Pix2pix}& \multirow{4}{*}{Cityscapes} & Original & 11.4M & -- & 56.8G & -- & \multicolumn{2}{c}{--} & 42.06 & -- \\
& & Ours (w/o fine-tuning) & 0.71M & 16.0$\times$ & 5.66G & 10.0$\times$ & \multicolumn{2}{c}{--} & 33.35 & 8.71\\
& & Ours & \textbf{0.71M} & \textbf{16.0$\times$} & 5.66G & 10.0$\times$ & \multicolumn{2}{c}{--} & 40.77 & 1.29 \\
& & Ours (Fast) & 0.89M & 12.8$\times$ & 5.45G & \textbf{10.4$\times$} & \multicolumn{2}{c}{--} & \textbf{41.71} & \textbf{0.35} \\
\cmidrule{2-11}
\multirow{4}{*}{} & \multirow{4}{*}{map$\to$arial photo} & Original & 11.4M & -- & 56.8G & -- & 47.76 & -- & \multicolumn{2}{c}{--} \\
& & Ours (w/o fine-tuning) & 0.75M & 15.1$\times$ & 4.68G & 12.1$\times$ & 71.82 & 24.1 & \multicolumn{2}{c}{--} \\
& & Ours & 0.75M & 15.1$\times$ & 4.68G & 12.1$\times$ & \textbf{48.02} & \textbf{0.26} & \multicolumn{2}{c}{--} \\
& & Ours (Fast) & \textbf{0.71M} & \textbf{16.1$\times$} & \textbf{4.57G} & \textbf{12.4$\times$} & 48.67 & 0.91 & \multicolumn{2}{c}{--} \\
\midrule
\multirow{7}{*}{GauGAN} & \multirow{4}{*}{Cityscapes} & Original & 93.0M & -- & 281G & -- & \multicolumn{2}{c}{--} & 62.18 & --\\
& & Ours (w/o fine-tuning) & 20.4M & 4.6$\times$ & 31.7G & 8.8$\times$ & \multicolumn{2}{c}{--} & 59.44 & 2.74 \\
& & Ours & 20.4M & 4.6$\times$ & 31.7G & 8.8$\times$ & \multicolumn{2}{c}{--} & \textbf{61.22} & 0.96 \\
& & Ours (Fast) & \textbf{20.2M} & \textbf{4.6}$\times$ & \textbf{31.2G} & \textbf{9.0}$\times$ & \multicolumn{2}{c}{--} & 61.17 & 1.01\\
\cmidrule{2-11}
& \multirow{3}{*}{COCO-Stuff} & Original & 97.5M & -- & 191G & -- & 21.96 & -- & 38.38 & --\\
& & Ours (Fast, w/o fine-tuning) & 26.0M & 3.8$\times$ & 35.4G & 5.4$\times$ & 28.78 & 6.82 & 31.77 & 6.61 \\
& & Ours (Fast) & \textbf{26.0M} & \textbf{3.8$\times$}& \textbf{35.4G} & \textbf{5.4}$\times$ & \textbf{25.06} & \textbf{3.10} & \textbf{35.34} & \textbf{3.04}  \\
\bottomrule
\end{tabular}
\caption{
Quantitative evaluation. We use the mIoU metric (the higher the better) for the Cityscapes and COCO-Stuff datasets, and FID (the lower the better) for other datasets. \textbf{Ours} denotes GAN Compression and \textbf{Ours (Fast)} denotes Fast GAN Compression as described in Section~\ref{sec:pipelines}. Our method can compress state-of-the-art conditional GANs by \textbf{9-30$\times$} in MACs and  \textbf{4-30$\times$} in model size, with only minor performance degradation.
For CycleGAN compression, our general-purpose approach outperforms previous CycleGAN-specific Co-Evolution method~\cite{shu2019co} by a large margin. Our faster and simpler pipeline, Fast GAN Compression, achieves similar results compared to GAN Compression. }

\label{tab:final}
\vspace{-10pt}
\end{table*}

After the once-for-all network is trained, we find the best-performing sub-network by \textit{directly} evaluating the performance of each candidate sub-network on the validation set through the brute-force search method under the certain computation budget $F$. Since the once-for-all network is thoroughly trained with weight sharing, no fine-tuning is needed. This approximates the model performance when it is trained from scratch. In this manner, we can decouple the training and search of the generator architecture: we only need to train once, but we can evaluate all the possible channel configurations without further training and pick the best one as the search result. Optionally, we fine-tune the selected architecture to further improve the performance. We report both variants in Section~\ref{sec:quantitative-results}.

\subsection{Fast GAN Compression}
\label{sec:pipelines}

\begin{figure}[t]
\centering
\begin{subfigure}[b]{\linewidth}
	\includegraphics[width=\textwidth]{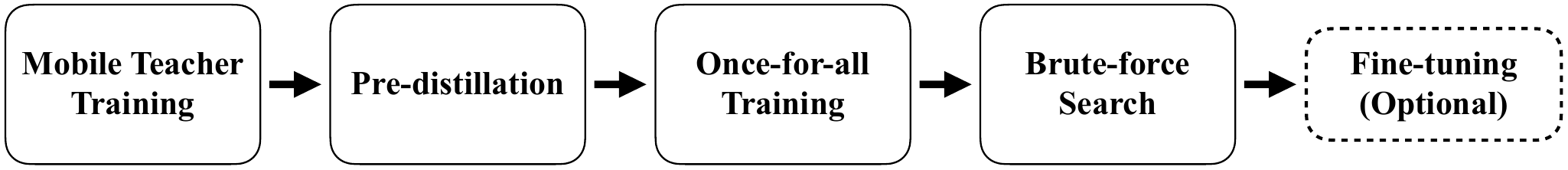}
	\caption{GAN Compression Pipeline}
	\label{fig:full_pipeline}
\end{subfigure}
\begin{subfigure}[b]{\linewidth}
	\includegraphics[width=\textwidth]{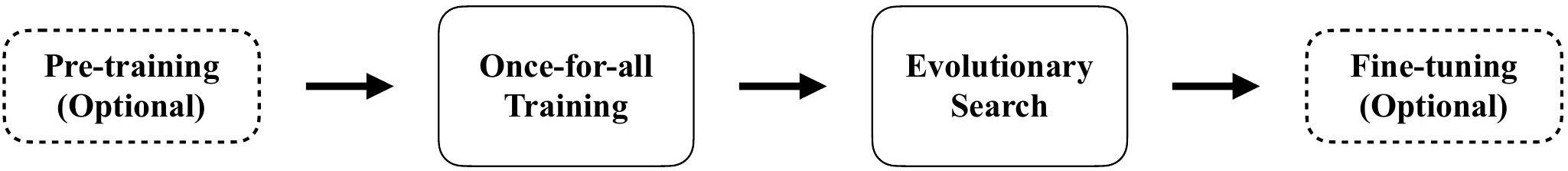}
	\caption{Fast GAN Compression Pipeline}
	\label{fig:lite_pipeline}
\end{subfigure}
\vspace{-15pt}
\caption{Pipelines of our GAN Compression and Fast GAN Compression. Fast GAN Compression does not require \emph{Mobile Teacher Training} and \emph{Pre-distillation}, and uses \emph{Evolutionary Search} instead of \emph{Brute-force Search}. The steps with dashed borders are optional. If the original model is available, we can also skip the \emph{Pre-training}. The \emph{Fine-tuning} step is also optional, as reported in Section~\ref{sec:decouple}.}
\label{fig:pipelines}
\vspace{-10pt}
\end{figure}

Although GAN Compression can accelerate cGAN generators significantly without losing visual fidelity, the whole training pipeline is both involved and slow. In this pipeline, for each model and dataset, we first train a MobileNet-style teacher network~\cite{howard2017mobilenets} from scratch and then pre-distill the knowledge of this teacher network to a student network.
Next, we train the once-for-all network by adopting the weights of distilled student network. 
We then evaluate all sub-networks under the computation budget $F$ with the brute-force search method. 
As the once-for-all network is initialized with the pre-distilled student network, the largest sub-network within the once-for-all network is exactly the pre-distilled student.
Therefore, pre-distillation helps reduce the search space by excluding the large sub-network candidates. 
After evaluation, we choose the best-performing sub-network within the once-for-all network and optionally fine-tune it to obtain our final compressed model. The detailed pipeline is shown in Figure~\ref{fig:full_pipeline}. Since we have to conduct \emph{Mobile Teacher Training}, \emph{Pre-distillation}, and \emph{Once-for-all Training} in this pipeline, the training cost is about 3-4$\times$ larger than the original model training. Besides, the brute-force search strategy has to evaluate \textit{all} candidate sub-networks, which is time-consuming and also limits the size of the search space. If not specified, our models are compressed with this full pipeline (named GAN Compression) in Section~\ref{sec:experiments}.

To handle the above issues, we propose an improved pipeline, Fast GAN Compression. This simpler and faster pipeline can produce comparable results as GAN Compression. In the training stage, we no longer need to train a MobileNet-style teacher network~\cite{howard2017mobilenets} and run the pre-distillation. Instead, we directly train a MobileNet-style once-for-all network~\cite{cai2020once} from scratch using the original full network as a teacher. In the search stage, instead of evaluating all sub-networks, we use the evolutionary algorithm~\cite{real2019regularized} to search for the best-performing sub-network under the computation budget $F$ within the once-for-all network. The evolutionary algorithm enables us to search in a larger space and makes the pre-distillation no longer needed. 
We keep a population of $P$ sub-networks throughout the whole search process. The population is initialized with models of  randomly sampled architectures that meet the computation budget $F$. After this, we run the following steps for $T$ iterations. At each iteration, the $r_p \cdot P$ best sub-networks form the parent group. Then we would generate $r_m \cdot P$ mutated samples and $(1-r_m)\cdot P$ crossovered samples by the parents as the next generation. Specifically, 
\begin{itemize}
    \item For mutation, we first uniformly sample an architecture configuration $\{c_k\}_{k=1}^K$ from parents. Each $c_k$ has a probability $p_m$ to mutate to other candidate values independently. If the mutated configuration does not meet the budget, we would re-mutate it until it meets the budget.
    \item For crossover, we first uniformly sample two architecture configurations $\{c^1_k\}_{k=1}^K$ and $\{c^2_k\}_{k=1}^K$ from parents. The the crossovered sample will be $\{c'_k\}_{k=1}^K$ where $c'_k$ is uniformly chosen from $\{c^1_k, c^2_k\}$. Also, if the new sample does not meet budget, we would resample it until it is under the budget.
\end{itemize}
The best-performed sub-network of all generations is the one we are looking for. 
After finding this best sub-network, we would also optionally fine-tune it with the pre-trained discriminator to obtain our final compressed model. 
The detailed differences between GAN Compression and Fast GAN Compression are shown in Figure~\ref{fig:pipelines}. 
With the optimized pipeline, we could save up to 70\% training time and 90\% search time of GAN Compression (see Section~\ref{sec:train_search_time}). 
Since the efficient search method in Fast GAN Compression allows a larger search space compared to GAN Compression (Section~\ref{sec:search_space}), the new pipeline achieves similar performance without additional steps.

\section{Experiments}
\label{sec:experiments}

\subsection{Setups}
\label{sec:setups}

\subsubsection{Models}
We conduct experiments on four conditional GAN models to demonstrate the generality of our method.
\begin{itemize}[leftmargin=*]
    \item CycleGAN~\cite{zhu2017unpaired}, an unpaired image-to-image translation model, uses a ResNet-based generator~\cite{he2016deep,johnson2016perceptual} to transform an image from a source domain to a target domain without using pairs. 
    \item Pix2pix~\cite{isola2017image} is a conditional-GAN-based paired image-to-image translation model. For this model, we replace the original U-Net generator~\cite{ronneberger2015u} with the ResNet-based generator~\cite{johnson2016perceptual} as we observe that the ResNet-based generator achieves better results with less computation cost, given the same learning objective. %See Section~\ref{sec:architecture} for a detailed U-Net \vs ResNet benchmark. 
    \item GauGAN~\cite{park2019semantic} is a recent paired image-to-image translation model. It can generate a high-fidelity image given a semantic label map.
    \item MUNIT~\cite{huang2018multimodal} is a multimodal unpaired image-to-image translation framework based on the encoder-decoder framework. It allows a user to control the style of  output images by providing a reference style image. Different from the single generator used in Pix2pix, CycleGAN, and GauGAN, MNUIT architecture includes three networks:  a style encoder, a content encoder, and a decoder network.
\end{itemize}
We retrained the Pix2pix and CycleGAN using the official PyTorch repo with the above modifications. Our retrained models (available at our \href{https://github.com/mit-han-lab/gan-compression}{repo}) slightly outperform the official pre-trained models. We use these retrained models as original models. For GauGAN and MUNIT, we use the pre-trained models from the authors. 
\renewcommand \arraystretch{0.95}
\begin{table}[!t]
\setlength{\tabcolsep}{2.3pt}
\small\centering
\begin{tabular}{cccc}
\toprule
\multirow{2}{*}[\multirowcenter]{Model} & \multirow{2}{*}[\multirowcenter]{Dataset} & \multicolumn{2}{c}{\#Sub-networks}
 \\ \cmidrule(lr){3-4} 
  &  & GAN Comp. & Fast GAN Comp. \\
\midrule
CycleGAN & horse$\to$zebra &
$6.6 \times 10^3$&
$3.9 \times 10^5$ \\
\midrule
MUNIT & edges$\to$shoes &
$9.8 \times 10^6$&
$2.8 \times 10^8$ \\
\midrule
& edges$\to$shoes &
$5.2 \times 10^3$&
$5.8 \times 10^6$\\
\cmidrule(lr){2-4}
Pix2pix & cityscapes &
$5.2 \times 10^3$&
$5.8 \times 10^6$\\
\cmidrule(lr){2-4}
& map$\to$arial photo & 
$5.2 \times 10^3$&
$5.8 \times 10^6$\\
\midrule
\multirow{2}{*}[\multirowcenter]{GauGAN} & Cityscapes &
$1.6 \times 10^4$& 
$1.3 \times 10^5$\\
\cmidrule(lr){2-4}
& COCO-Stuff & -- &
$3.9\times 10^5$\\
\bottomrule
\end{tabular}
\caption{The detailed search space sizes of \textbf{GAN Compression~(GAN Comp.)} and \textbf{Fast GAN Compression~(Fast GAN Comp.)}. We take the number of sub-networks within a search space as the size of the search space. For the newest experiment of GauGAN on COCO-Stuff, we directly apply Fast GAN Compression, so we do not have the search space for the GAN Compression. Benefiting from removing \textit{pre-distillation} and the evolutionary search algorithm, Fast GAN Compression could support a much larger search space than GAN Compression.}
\label{tab:search_space}
\vspace{-10pt}
\end{table}

\subsubsection{Datasets}
We use the following \ndatasets datasets in our experiments:
\begin{itemize}[leftmargin=*]
    \item Edges$\to$shoes. We use 50,025 images from UT Zappos50K dataset~\cite{yu2014fine}. We split the dataset randomly so that the validation set has 2,048 images for a stable evaluation of \fid (FID)~(see Section~\ref{subsec:metrics}). We evaluate the Pix2pix and MUNIT models on this dataset. \label{sec:edges2shoes}
    \item Cityscapes. The dataset~\cite{cordts2016cityscapes} contains the images of German street scenes. The training set and the validation set consists of 2975 and 500 images, respectively. We evaluate both the Pix2pix and GauGAN models on this dataset.
    \item Horse$\leftrightarrow$zebra. The dataset consists of 1,187 horse images and 1,474 zebra images originally from ImageNet~\cite{deng2009imagenet} and is used in CycleGAN~\cite{zhu2017unpaired}. The validation set contains 120 horse images and 140 zebra images. We evaluate the CycleGAN model on this dataset.
    \item Map$\leftrightarrow$aerial photo. The dataset contains 2194 images scraped from Google Maps and used in pix2pix~\cite{isola2017image}. The training set and the validation set contains 1096 and 1098 images, respectively. We evaluate the pix2pix model on this dataset.
    \item COCO-Stuff. COCO-Stuff~\cite{caesar2018COCO} dataset is derived from the COCO dataset~\cite{lin2014microsoft} with 118,000 training images and 5,000 validation images. We evaluate the GauGAN model on this dataset.
\end{itemize}

\subsubsection{Search Space}
\label{sec:search_space}
The search space design is critical for once-for-all~\cite{cai2020once} network training. Generally, a larger search space will produce more efficient models.
We remove certain channel number options in earlier layers to reduce the search cost, such as the input feature extractors. In the original GAN Compression pipeline, we load the pre-distilled student weights for the once-for-all training. As a result,   the pre-distilled network is the largest sub-network and much smaller than the mobile teacher. This design excludes many large sub-network candidates and may hurt the performance. In Fast GAN Compression, we do not have such constraint as we directly training a once-for-all network of the mobile teacher's size. Also, Fast GAN Compression uses a much more efficient search method, so its search space can be larger.
The detailed search space sizes of GAN Compression and Fast GAN Compression are shown in Table~\ref{tab:search_space}. Please refer to our \href{https://github.com/mit-han-lab/gan-compression}{code} for more details about the search space of candidate sub-networks.
 
\subsubsection{Implementation Details}
For the CycleGAN and Pix2pix models, we use the learning rate of 0.0002 for both generator and discriminator during training in all the experiments. The batch sizes on dataset horse$\to$zebra, edges$\to$shoes, map$\to$aerial photo, and cityscapes are 1, 4, 1, and 1, respectively. For the GauGAN model, we follow the setting in the original paper~\cite{park2019semantic}, except that the batch size is 16 instead of 32. We find that we can achieve a better result with a smaller batch size. 
For the evolutionary algorithm, the population size $P=100$, the number of iterations $T=500$, parent ratio $r_p=0.25$, mutation ratio $r_m=0.5$, and mutation probability $r_p=0.2$.
Please refer to our \href{https://github.com/mit-han-lab/gan-compression}{code} for more implementation details.

\renewcommand \arraystretch{1.}
\begin{table}[t]
\setlength{\tabcolsep}{0pt}
\small\centering
\begin{tabular}{lccccc}
\toprule
Model & Dataset & Method & Preference & \multicolumn{2}{c}{LPIPS ($\downarrow$)} \\
\midrule
\multirow{6}{*}{Pix2pix}& \multirow{3}{*}{edges$\to$shoes} & Original& -- & \multicolumn{2}{c}{0.185} \\
& &  Ours & -- & 0.193 & \textbf{(-0.008)} \\
& & 0.28 Pix2pix & -- & 0.201 & (-0.016)\\ 
\cmidrule(lr){2-6}
& \multirow{3}{*}{cityscapes} & Original& -- & \multicolumn{2}{c}{0.435} \\
& &  Ours & -- & 0.436 & \textbf{(-0.001)} \\
& & 0.31 Pix2pix & -- & 0.442 & (-0.007)\\ 
\midrule
\multirow{2}{*}{CycleGAN}& \multirow{2}{*}{horse$\to$zebra} & Ours & 72.4\% & \multicolumn{2}{c}{--} \\
& &  0.25 CycleGAN & 27.6\% & \multicolumn{2}{c}{--} \\  
\bottomrule
\end{tabular}
\caption{Perceptual study: The LPIPS~\cite{zhang2018unreasonable} is a perceptual metric for evaluating the similarity between a generated image and its corresponding ground-truth real image. The lower LPIPS indicates better perceptual photorealism of the results. This reference-based metric requires paired data. For the unpaired setting, such as the horse$\to$zebra dataset, we conduct a human study for our GAN Compression method and the 0.25 CycleGAN. We ask human participants which generated image looks more like a zebra. 72.4\% workers favor results from our model.} 
\label{tab:perceptual}
\vspace{-15pt}
\end{table}

\subsubsection{Intermediate Distillation Layers}
\label{subsubsec:distillation layers}
For the ResNet generator in CycleGAN and Pix2pix, we choose 4 intermediate activations for distillation. We split the 9 residual blocks into groups of size 3 and use feature distillation every three layers. We empirically find that such a configuration can transfer enough knowledge while it is easy for the student network to learn, as shown in Section~\ref{sec:distillation}.

For the GauGAN generator with 7 SPADE ResNetBlocks, we choose the output of the first, third, and fifth SPADE ResNetBlock for intermediate feature distillation.

For the multimodal encoder-decoder models such as MUNIT, we choose to use the style code (i.e., the output of the style encoder) and the content code (i.e., the output of the content encoder) for intermediate feature distillation. Matching the style code helps preserve the multimodal information stored in the style encoder while matching the content code further improves the visual fidelity as the content code contains high-frequency information of the input image.

\subsubsection{MobileNet-Style Architectures}
\label{subsubsec:mobile arch}
For the ResNet generator in CycleGAN and Pix2pix, we decompose all the convolutions in the ResBlock layers, which consume more than 75\% computation of the generator.

For the GauGAN generator, we decompose the $\beta$ and $\gamma$ convolutions in the SPADE modules. The SPADE modules account for about 80\% computation of the generator, and $\beta$ and $\gamma$ convolutions consumes about 88\% computation of the SPADE modules, which are highly redundant.

For the MUNIT generator, we decompose the ResBlock layers in the content encoder and decoder, and the upsample layers in the decoder. These modules consume the majority of the model computation. The style encoder remains unchanged since it only consumes less than 10\% of computation.

\begin{figure*}[!h]
\centering
\vspace{-15pt}
\includegraphics[width=\linewidth]{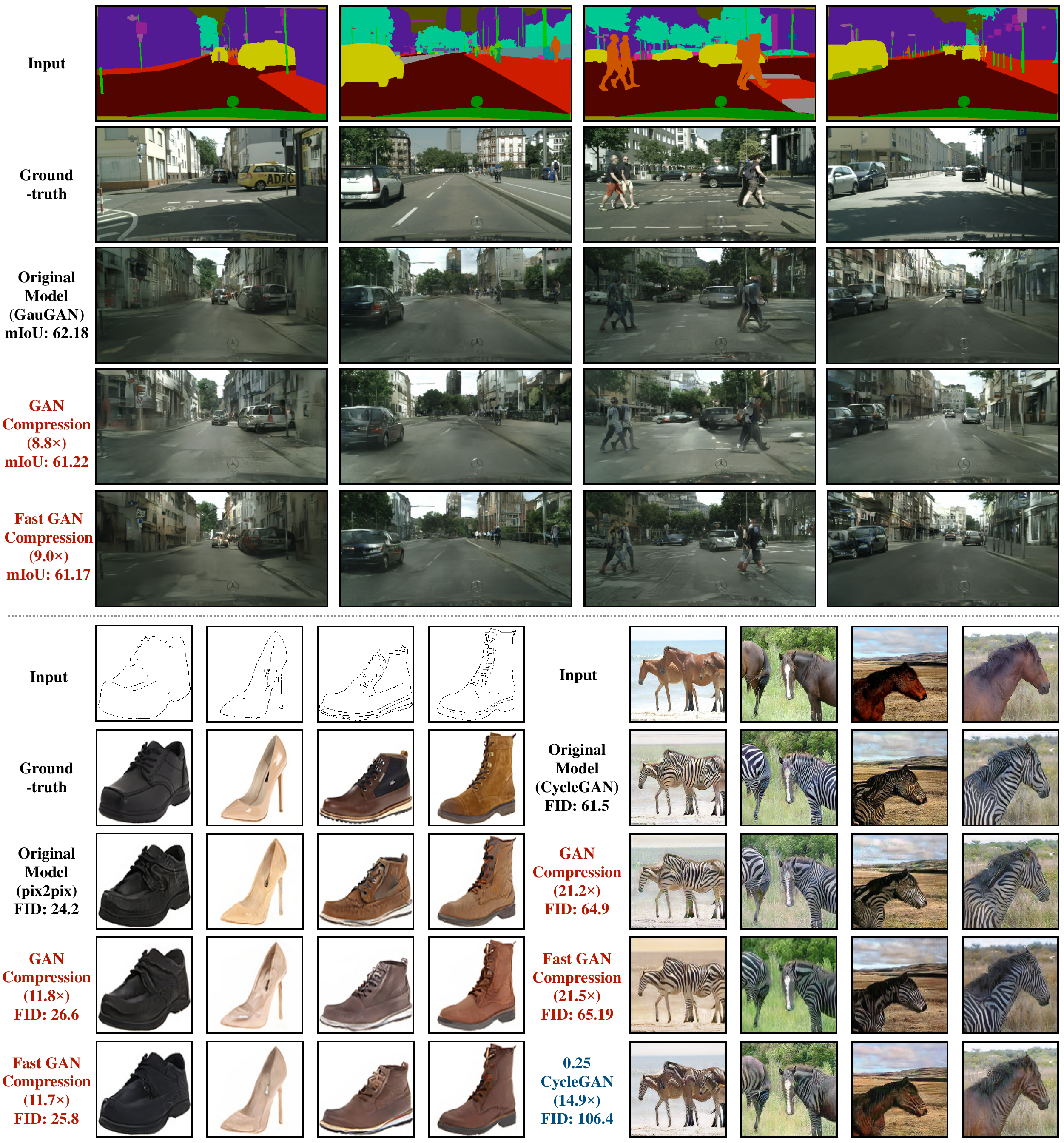}
\caption{Qualitative compression results on Cityscapes, edges$\rightarrow$shoes, and horse$\rightarrow$zebra. Our methods (GAN Compression and Fast GAN Compression) preserve the fidelity while significantly reducing the computation. In contrast, directly training a smaller model (\eg, 0.25 CycleGAN, which linearly scales each layer to 25\% channels) yields poor performance.}
\label{fig:qualitative}
\vspace{-10pt}
\end{figure*}

\begin{figure*}[h]
\centering
\vspace{-10pt}
\includegraphics[width=0.9\linewidth]{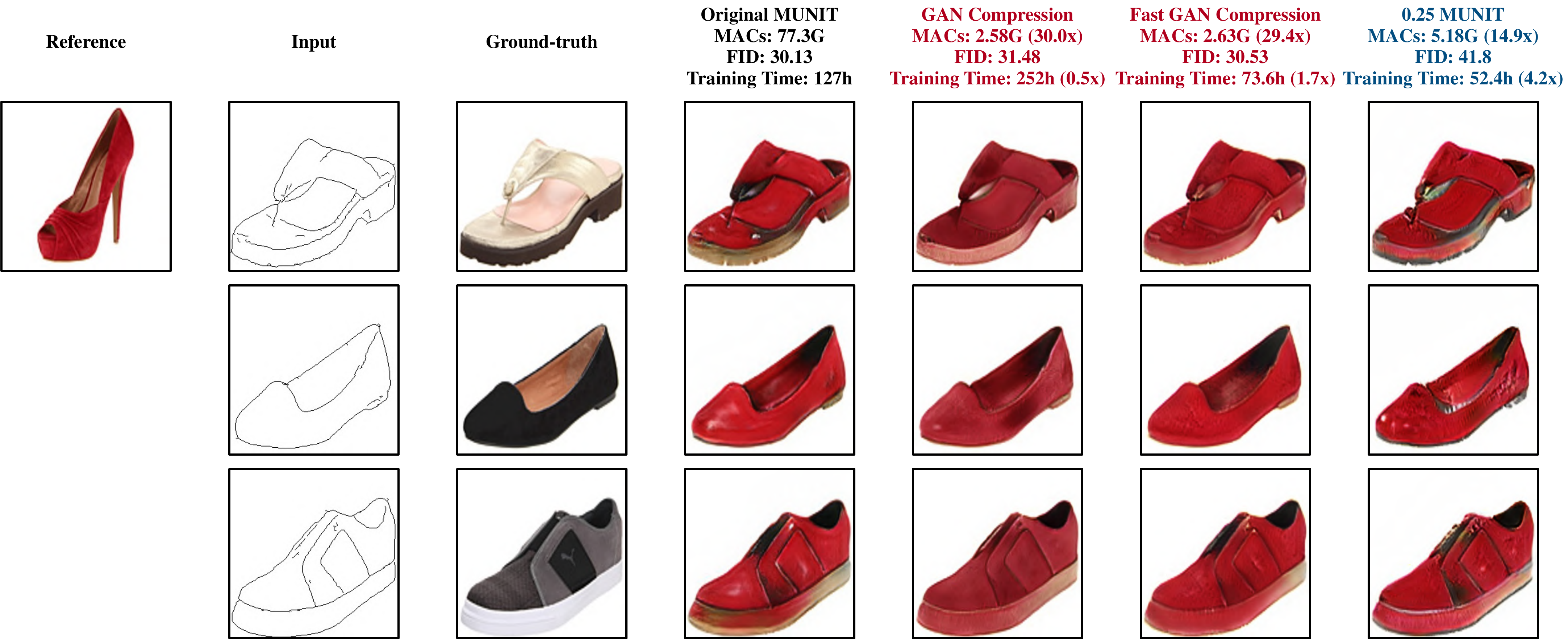}
\caption{Qualitative compression results on edges$\to$shoes of MUNIT. Both GAN Compression and Fast GAN Compression could preserve the style of the reference image and the visual fidelity while reducing the computation significantly, but Fast GAN Compression only needs 30\% training time. On the contrary, directly training a smaller model (0.25 MUNIT, which means linearly scales each layer to 25\% channels) yields poor performance.} 
\label{fig:edges-munit-ref}
\vspace{-10pt}
\end{figure*}

\subsection{Evaluation Metrics}
\label{subsec:metrics}
We introduce the metrics for assessing the equality of synthesized images.

\subsubsection{\fid (FID)~\cite{heusel2017gans}}
The FID score aims to calculate the distance between the distribution of feature vectors extracted from real and generated images using an InceptionV3~\cite{szegedy2016rethinking} network. The score measures the similarity between the distributions of real and generated images. A \emph{lower} score indicates a \emph{better} quality of generated images. We use an open-sourced FID evaluation code\footnote{\url{https://github.com/mseitzer/pytorch-fid}}. For paired image-to-image translation (pix2pix and GauGAN), we calculate the FID between translated test images to real test images. For unpaired image-to-image translations (CycleGAN), we calculate the FID between translated test images to real training+test images. This allows us to use more images for a stable FID evaluation, as done in previous unconditional GANs~\cite{karras2019style}. The FID of our compressed CycleGAN model slightly increases when we use real test images instead of real training+test images.

\subsubsection{Semantic Segmentation Metrics}
Following prior work~\cite{isola2017image,zhu2017unpaired,park2019semantic}, we adopt a semantic segmentation metric to evaluate the generated images on the Cityscapes and COCO-Stuff dataset. We run a semantic segmentation model on the generated images and compare how well the segmentation model performs. 
We choose the mean Intersection over Union~(mIoU) as the segmentation metric, and we use DRN-D-105~\cite{yu2017dilated} as our segmentation model for Cityscapes and DeepLabV2~\cite{chen2017deeplab} for COCO-Stuff. \emph{Higher} mIoUs suggest that the generated images look more \emph{realistic} and better reflect the input label map. For Cityscapes, we upsample the DRN-D-105's output semantic map to 2048$\times$1024, which is the resolution of the ground truth images. For COCO-Stuff, we resize the generated images to the resolution of the ground truth images. Please refer to our \href{https://github.com/mit-han-lab/gan-compression}{code} for more evaluation details.

\subsection{Results}

\subsubsection{Quantitative Results}
\label{sec:quantitative-results}
We report the quantitative results of compressing CycleGAN, Pix2pix, GauGAN, and MUNIT on \ndatasets datasets in Table~\ref{tab:final}. Here we choose a small computation budget so that our compressed models could still largely preserve the visual fidelity under the budget. By using the best sub-network from the once-for-all network, our methods (GAN Compression and Fast GAN Compression) achieve large compression ratios. It can compress recent conditional GANs by \textbf{9-30$\times$} and reduce the model size by \textbf{4-33$\times$}, with only negligible degradation in the model performance. Specifically, our proposed methods show a clear advantage of CycleGAN compression compared to the previous Co-Evolution method~\cite{shu2019co}. We can reduce the computation of the CycleGAN generator by 21.2$\times$, which is 5$\times$ better compared to the previous CycleGAN-specific method~\cite{shu2019co}, while achieving a better FID by more than $30$\footnote{In CycleGAN setting, for our model, the original model, and baselines, we report the FID between translated test images and real training+test images, while Shu \etal~\cite{shu2019co}'s FID is calculated between translated test images and real test images. The FID difference between the two protocols is small. The FIDs for the original model, Shu \etal~\cite{shu2019co}, and our compressed model are 65.48, 96.15, and 69.54 using their protocol.}. Besides, the results produced by Fast GAN Compression are also on par with GAN Compression.

\subsubsection{Perceptual Similarity and User Study}
For the paired datasets such as edges$\to$shoes and Cityscapes, we evaluate the perceptual photorealism of our results. We use the LPIPS metric~\cite{zhang2018unreasonable}  to measure the perceptual similarity of generated images and the corresponding real images. A lower LPIPS indicates a better reconstruction of the generated images. For the CycleGAN model, we conduct a human preference test on the horse$\to$zebra dataset on Amazon Mechanical Turk (AMT) as there are no paired images. We basically follow the protocol of Isola~\etal~\cite{isola2017image}, except that we ask the workers to decide which image is more like a real zebra image between our GAN Compression model and 0.25 CycleGAN. Table \ref{tab:perceptual} shows our perceptual study results on both Pix2pix and CycleGAN. Our GAN Compression method significantly outperforms the straightforward  training baselines with a reduced number of channels.

\begin{figure}[!t]
\centering
\begin{subfigure}[b]{0.235\textwidth}
	\includegraphics[height=0.95\textwidth]{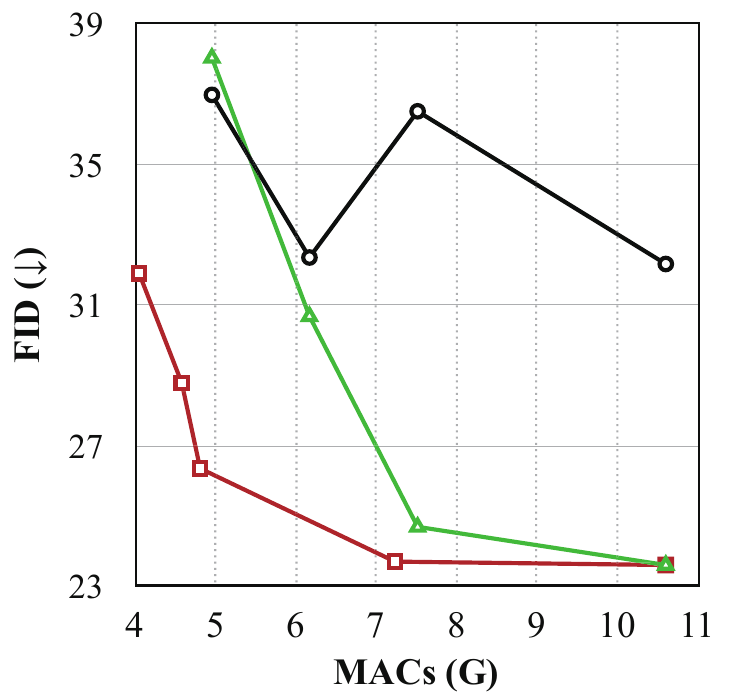}
	\caption{Edge $\to$ shoes.}
	\label{fig:tradeoff_edge}
\end{subfigure}
\begin{subfigure}[b]{0.235\textwidth}
    \includegraphics[height=0.95\textwidth]{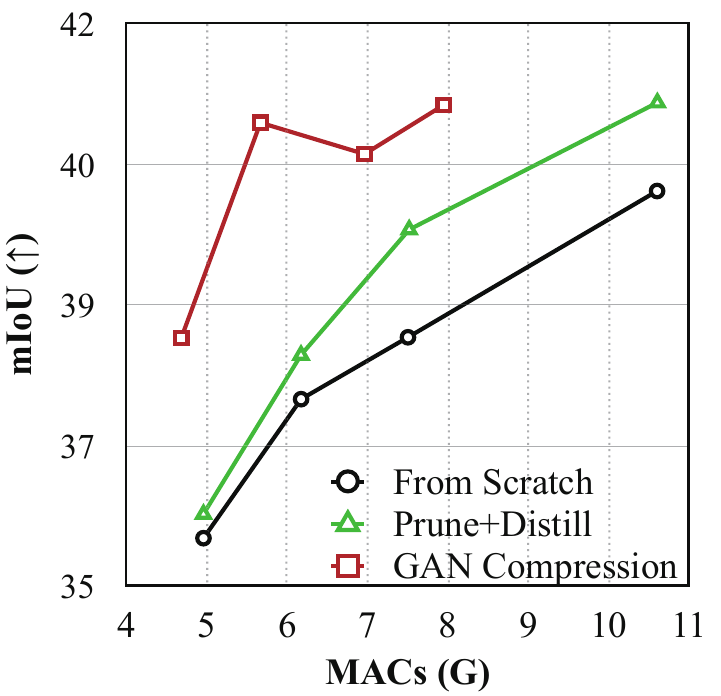}
	\caption{Cityscapes.}
	\label{fig:tradeoff_cityscapes}
\end{subfigure}
\caption{Trade-off curve of pix2pix on Cityscapes and edges$\to$shoes dataset. The Prune+Distill method outperforms training from scratch for larger models but works poorly when the model is aggressively shrunk. Our GAN Compression method can consistently improve the performance \vs computation trade-off at various scales.}
\label{fig:pareto}
\vspace{-10pt}
\end{figure}

\subsubsection{Qualitative Results}
Figure~\ref{fig:qualitative} shows several example results of GauGAN, Pix2pix, and CycleGAN, and Figure~\ref{fig:edges-munit-ref} shows the results of MUNIT given a specific style reference image. We provide the input, its ground truth (except for unpaired setting), the output of the original model, and the output of our compressed models. For the MUNIT, we also include the style reference image. Both of our compression methods closely preserve the visual fidelity of the output image even under a large compression ratio. For CycleGAN and MUNIT, we also provide the output of a baseline model (0.25 CycleGAN/MUNIT). The baselines 0.25 CycleGAN/MUNIT contain 25\% channels and are trained from scratch. Our advantage is clear: the baseline CycleGAN model can hardly synthesize zebra stripes on the output image, and the images generated by the baseline MUNIT model have some notable artifacts, given a much smaller compression ratio. There might be some cases where our compressed models show a small degradation (\eg,  the leg of the second zebra in Figure~\ref{fig:qualitative}), but our compressed models sometimes surpass the original one in other cases (\eg, the first and last shoe images have a better leather texture). Generally, GAN models compressed by our methods perform comparatively with respect to the original model, as shown by quantitative results (Table~\ref{tab:final}).

\subsubsection{Performance \vs Computation Trade-off}
Apart from the large compression ratio we can obtain, we verify that our method can consistently improve the performance at different model sizes. Taking the Pix2pix model as an example, we plot the performance \vs computation trade-off on Cityscapes and Edges$\to$shoes dataset in Figure~\ref{fig:pareto}.

First, in the large model size regime, Prune+Distill (without NAS) outperforms training from scratch, showing the effectiveness of intermediate layer distillation. Unfortunately, with the channels continuing shrinking down uniformly, the capacity gap between the student and the teacher becomes too large. As a result, the knowledge from the teacher may be too recondite for the student, in which case the distillation may even have negative effects on the student model. 

On the contrary, our training strategy allows us to automatically find a sub-network with a smaller gap between the student and teacher model, which makes learning easier. Our method consistently outperforms the other two baselines by a large margin.

\renewcommand \arraystretch{1.}
\begin{table}[t]
\setlength{\tabcolsep}{0.9pt}
\small\centering
\begin{tabular}{cccccccc}
\toprule
\multicolumn{2}{c}{Model} & \multicolumn{2}{c}{CycleGAN} & \multicolumn{2}{c}{Pix2pix} & \multicolumn{2}{c}{GauGAN}  \\
\midrule
\multirow{2}{*}[\multirowcenter]{Metric} &FID ($\downarrow$) & \multicolumn{2}{c}{61.5$\to$65.0} & \multicolumn{2}{c}{24.2$\to$26.6} & \multicolumn{2}{c}{--}\\
\cmidrule{2-8}
&mIoU ($\uparrow$) & \multicolumn{2}{c}{--} & \multicolumn{2}{c}{--} & \multicolumn{2}{c}{62.2 $\to$ 61.2} \\
\midrule
\multicolumn{2}{c}{MAC Reduction} & \multicolumn{2}{c}{21.2$\times$} & \multicolumn{2}{c}{11.8$\times$} & \multicolumn{2}{c}{8.8$\times$} \\
\multicolumn{2}{c}{Memory Reduction} & \multicolumn{2}{c}{2.0$\times$} & \multicolumn{2}{c}{1.7$\times$} & \multicolumn{2}{c}{1.8$\times$} \\
\midrule
Xavier& CPU & 1.65s &(18.5$\times$) & 3.07s&(9.9$\times$)& 21.2s &(7.9$\times$)\\
\cmidrule{2-8}
Speedup& GPU & 0.026s&(3.1$\times$)& 0.035s&(2.4$\times$)& 0.10s &(3.2$\times$)\\
\midrule
Nano&CPU&6.30s&(14.0$\times$)&8.57s&(10.3$\times$)&65.3s&(8.6$\times$)\\
\cmidrule{2-8}
Speedup& GPU&0.16s&(4.0$\times$)&0.26s&(2.5$\times)$&0.81s&(3.3$\times$)\\
\midrule
\multicolumn{2}{c}{1080Ti Speedup}&0.005s&(2.5$\times$)&0.007s&(1.8$\times$)&0.034s&(1.7$\times$)\\
\midrule
\multicolumn{2}{c}{Xeon Silver 4114}&\multirow{2}{*}{0.11s}&\multirow{2}{*}{(3.4$\times$)}&\multirow{2}{*}{0.15s}&\multirow{2}{*}{(2.6$\times$)}&\multirow{2}{*}{0.74s}&\multirow{2}{*}{(2.8$\times)$} \\
\multicolumn{2}{c}{CPU Speedup} \\
\bottomrule
\end{tabular}
\caption{Measured memory reduction and latency speedup on NVIDIA Jetson AGX Xavier, NVIDIA Jetson Nano, 1080Ti GPU, and Xeon CPU. CycleGAN, pix2pix, and GauGAN models are trained on horse$\to$zebra, edges$\to$shoes and Cityscapes datasets. GAN Compression could bring significant speedup on various hardware devices.}
\vspace{-10pt}
\label{tab:runtime}
\end{table}
\begin{figure}[t]
\centering
\begin{subfigure}[b]{0.48\linewidth}
    	\includegraphics[height=0.99\textwidth]{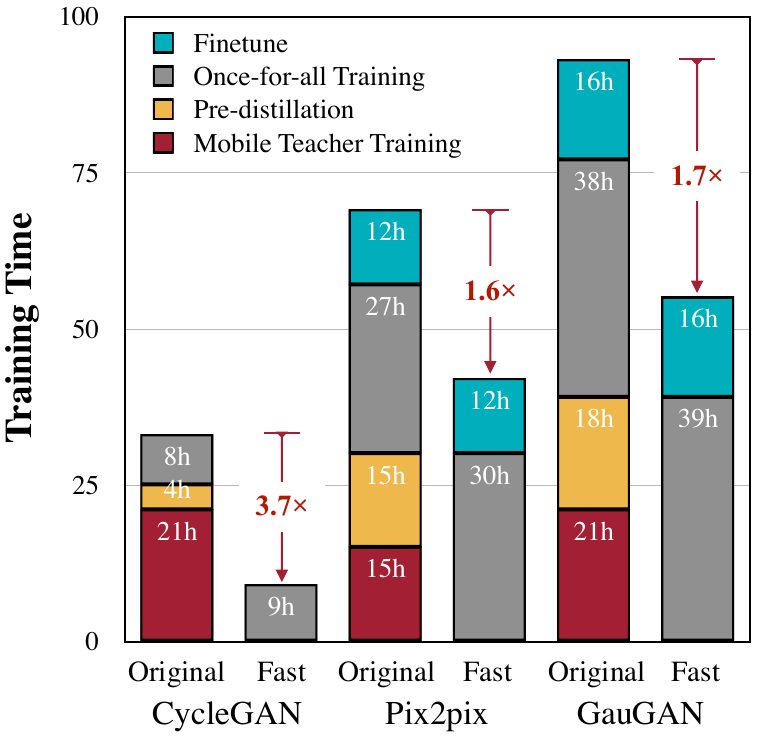}
	\caption{Training Time}
	\label{fig:train_time}
\end{subfigure}
\begin{subfigure}[b]{0.48\linewidth}
	\includegraphics[height=0.99\textwidth]{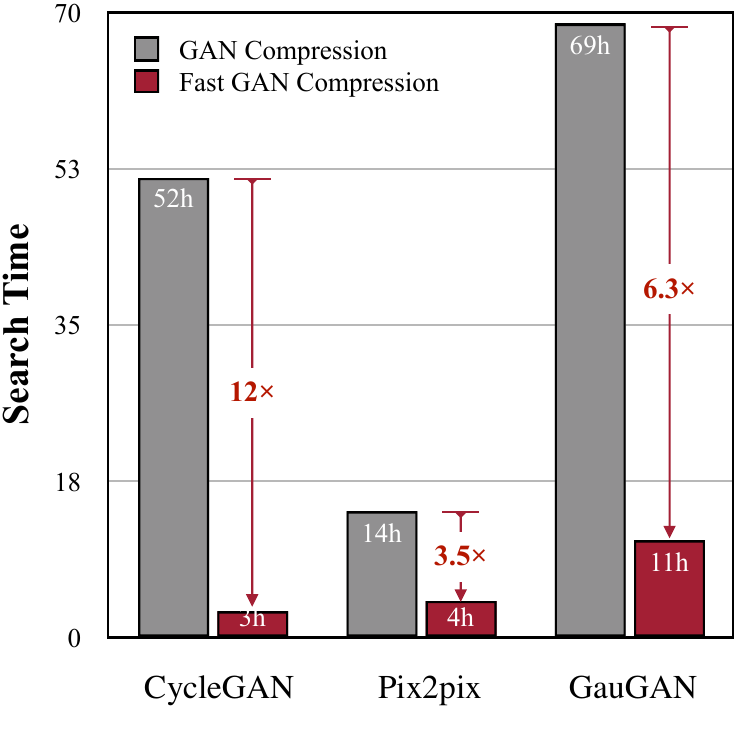}
	\caption{Search Time}
	\label{fig:search_time}
\end{subfigure}
\vspace{-5pt}
\caption{Training and search time of GAN Compression (\textbf{Original} in (a)) and Fast GAN Compression (\textbf{Fast} in (a)). Our Fast GAN Compression could save 1.7-3.7$\times$ training time and 3.5-12$\times$ search time. CycleGAN, Pix2pix, and GauGAN models are measured on horse$\to$zebra, edges$\to$shoes, and Cityscapes datasets, respectively. The training time of GauGAN is measured on 8 2080Ti GPUs, while others are all on a single 2080Ti GPU.}
\label{fig:train-search}
\vspace{-10pt}
\end{figure}

\begin{figure*}[h]
\centering
\vspace{-8pt}
\includegraphics[width=\linewidth]{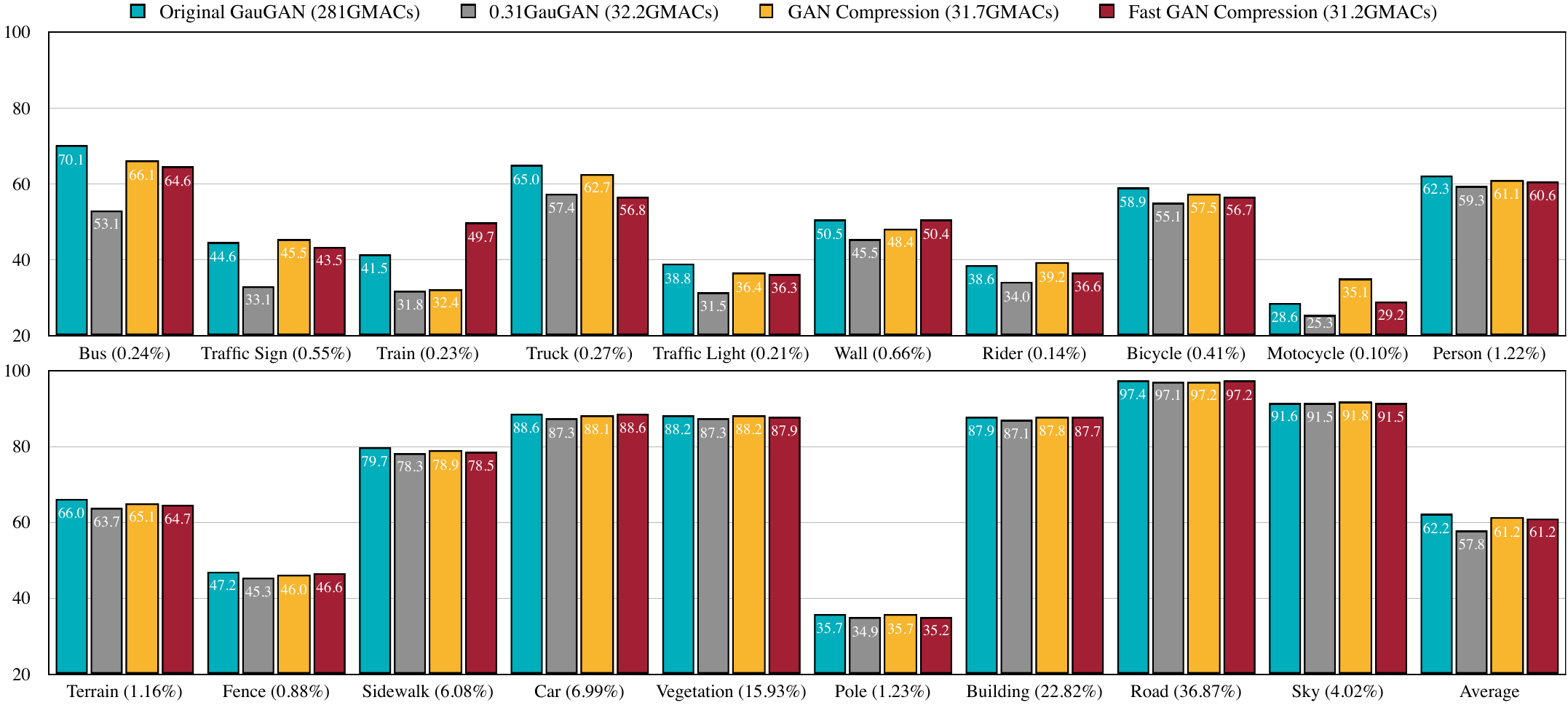}
\caption{Per-class IoU results of GauGAN compression on Cityscapes. The numbers in the brackets are the pixel frequency of this class in the training set. We sort the classes by the performance drop of the 0.31 GauGAN. Directly training a smaller model, 0.31GauGAN (i.e., linearly scales each layer to 31\% channels), hurts the IoU dramatically, especially for the rare categories (\eg, bus and traffic sign) compared to the dominant object categories (\eg, sky and road). Our methods can preserve the image quality effectively.}
\label{fig:cityscapes-ious}
\vspace{-10pt}
\end{figure*}

\subsubsection{Inference Acceleration on Hardware}
For real-world interactive applications, inference acceleration on hardware is more critical than the reduction of computation. To verify the practical effectiveness of our method, we measure the inference speed of our compressed models on several devices with different computational powers. To simulate interactive applications, we use a batch size of 1. We first perform 100 warm-up runs and measure the average timing of the next 100 runs. The results are shown in Table~\ref{tab:runtime}. 
The inference speed of the compressed CycleGAN generator on edge GPU of Jetson Xavier can achieve about \textbf{40} FPS, meeting the demand of interactive applications. We notice that the acceleration on GPU is less significant compared to CPU, mainly due to the large degree of parallelism. Nevertheless, we focus on making generative models more accessible on edge devices where powerful GPUs might not be available, so that more people can use interactive cGAN applications.

\begin{figure*}[!h]
\centering
\includegraphics[width=0.95\linewidth]{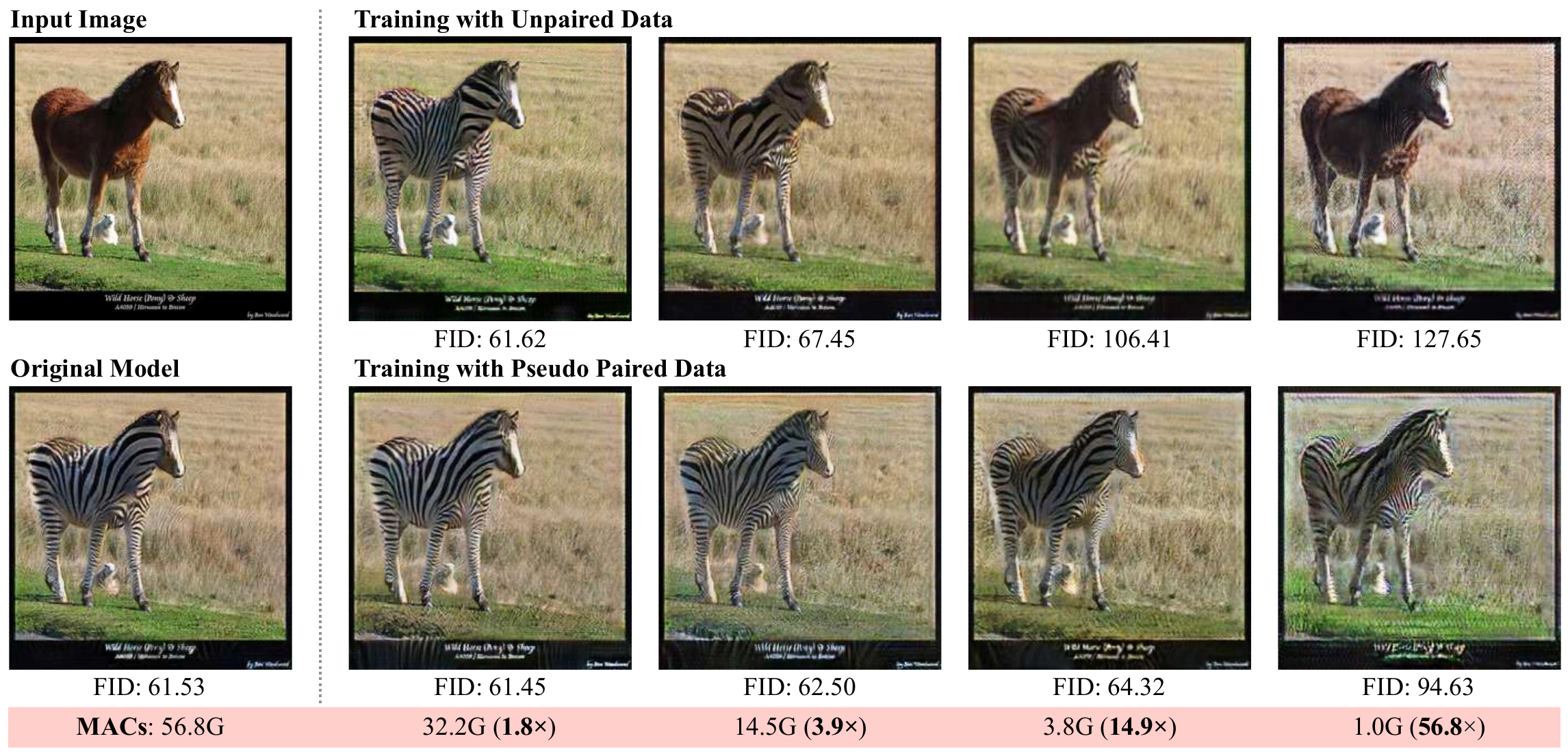}
\caption{The comparison between training with unpaired data (naive) and training with pseudo paired data (proposed).  The latter consistently outperforms the former, especially for small models. The generator's computation can be compressed by $14.9 \times$ without hurting the fidelity using the proposed pseudo pair method. In this comparison, both methods do not use automated channel reduction and convolution decomposition.}
\vspace{-15pt}
\label{fig:horse}
\end{figure*}
\begin{figure}[!h]
\centering
\begin{subfigure}[b]{0.235\textwidth}
    	\includegraphics[height=0.95\textwidth]{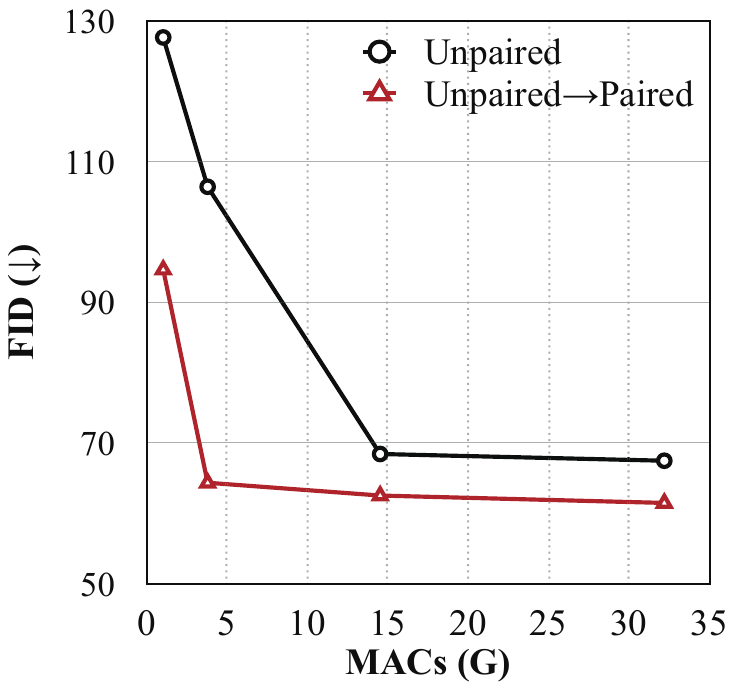}
	\caption{Unpaired \vs paired.}
	\label{fig:ablation_unpaired_paired}
\end{subfigure}
\begin{subfigure}[b]{0.235\textwidth}
	\includegraphics[height=0.95\textwidth]{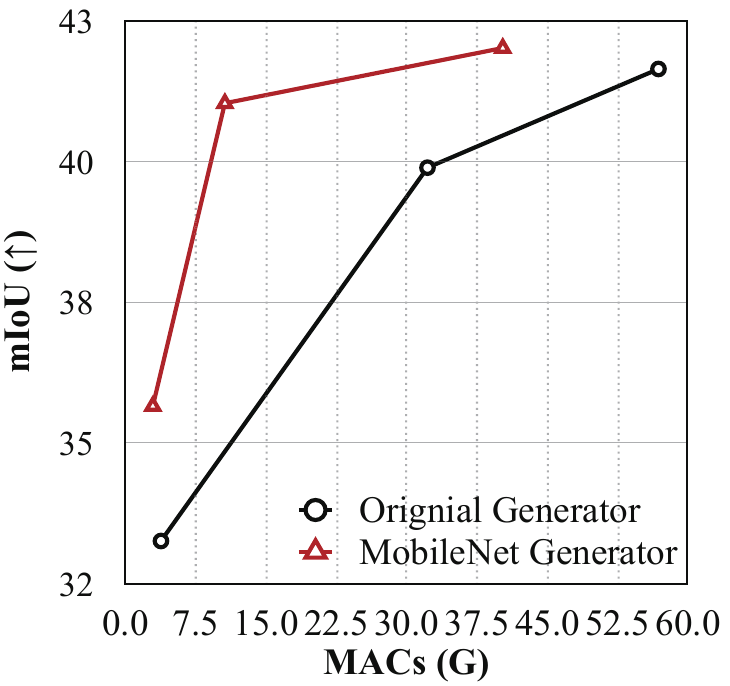}
	\caption{Normal \vs mobile conv.}
	\label{fig:ablation_original_mobile}
\end{subfigure}
\vspace{-5pt}
\caption{Ablation study: \textbf{(a)} Transforming the unpaired training into a paired training (using the pseudo pairs generated by the teacher model) significantly improves the performance of efficient models. \textbf{(b)} Decomposing the convolutions in the original ResNet-based generator into depth-wise and point-wise convolutional filters improves the performance \vs computation trade-off. We call our modified network MobileNet generator. }
\vspace{-5pt}
\end{figure}

\subsubsection{Training and Search Time}
\label{sec:train_search_time}
In Figure~\ref{fig:train-search}, we compare the training and search time of GAN Compression and Fast GAN Compression for CycleGAN, Pix2pix, and GauGAN on horse$\to$zebra, edges$\to$shoes, and Cityscapes dataset, respectively. After we remove the \emph{Mobile Teacher Training} and \emph{Pre-distillation} in Figure~\ref{fig:pipelines}, Fast GAN Compression accelerates the training of GAN Compression by  1.6-3.7$\times$, as shown in Figure~\ref{fig:train_time}. When switching to the evolutionary algorithm from the brute-force search, the search cost reduces dramatically, by 3.5-12$\times$, as shown in Figure~\ref{fig:search_time}.

\subsubsection{Cityscapes Per-class Performance}
We show the per-class IoU comparisons of GauGAN on Cityscapes dataset in Figure~\ref{fig:cityscapes-ious}. In general, the IoU of the dominant categories (\eg, sky and road in the bottom chart) are higher than those of the rare classes (\eg, bus and traffic sign in the top chart). When directly training a smaller model (0.31 GauGAN), the segmentation performance of these rare categories drops dramatically, while the dominant categories are less sensitive to the compression. With GAN Compression or Fast GAN Compression, we can significantly alleviate such performance drop.

\subsection{Ablation Study}
\label{sec:ablation}
Below we perform several ablation studies regarding our individual system components and design choices. 

\renewcommand \arraystretch{0.95}
\begin{table}[!t]
\setlength{\tabcolsep}{3.8pt}
\small\centering
\begin{tabular}{llccccc}
\toprule
\multirow{2}{*}[\multirowcenter]{Dataset} & \multirow{2}{*}[\multirowcenter]{Setting} & \multicolumn{3}{c}{Training Technique} & \multicolumn{2}{c}{Metric}  \\ \cmidrule(lr){3-5} \cmidrule(lr){6-7}
  &  & Pr. & Dstl. & Keep D. & FID ($\downarrow$) & mIoU ($\uparrow$) \\
\midrule
\multirow{10}{*}[\multirowcenter]{
\shortstack{Edges\\$\to$ \\ shoes}
} & ngf=64  & & & & \textbf{24.91}&-- \\  
\cmidrule(lr){2-7}
& \multirow{8}{*}[\multirowcenter]{ngf=48}& & & & 27.91 &-- \\ \cmidrule(lr){3-7}
&  & & &$\checkmark$ & 28.60&-- \\ \cmidrule(lr){3-7}
&  & $\checkmark$ &  & $\checkmark$&  27.25&-- \\ \cmidrule(lr){3-7}
&  & & $\checkmark$ & $\checkmark$&  26.32&-- \\ \cmidrule(lr){3-7}
&  & $\checkmark$ & $\checkmark$ & & 46.24&-- \\ \cmidrule(lr){3-7}
&  & $\checkmark$ & $\checkmark$ & $\checkmark$ & \textbf{24.45}&-- \\ 
\midrule
& ngf=96  & & &  & --& \textbf{42.47} \\ \cmidrule(lr){2-7}
& \multirow{8}{*}[\multirowcenter]{ngf=64}& & & & --& 40.49 \\ \cmidrule(lr){3-7}
&  & & &  $\checkmark$& --& 38.64 \\  \cmidrule(lr){3-7}
Cityscapes&  & $\checkmark$ & & $\checkmark$ & --&  40.98 \\  \cmidrule(lr){3-7}
&&  & $\checkmark$ &  $\checkmark$& --& 41.49 \\  \cmidrule(lr){3-7}
&  &  $\checkmark$ &  $\checkmark$ & & --&  40.66 \\  \cmidrule(lr){3-7}
&  &  $\checkmark$ & $\checkmark$ & $\checkmark$ &   --&  \textbf{42.11} \\
\bottomrule
\end{tabular}
\caption{Ablation study. \textbf{Pr.}: Pruning; \textbf{Dstl}: Distillation; \textbf{Keep D.}: in this setting, we inherit the discriminator's weights from the teacher discriminator. Pruning combined with distillation achieves the best performance on both datasets.}
\label{tab:ablation}
\vspace{-5pt}
\end{table}

\subsubsection{Advantage of Unpaired-to-paired Transform} 
\label{sec:unpaired-to-paired}

We first analyze the advantage of transforming unpaired conditional GANs into a pseudo paired training setting using the teacher model's output.

Figure~\ref{fig:horse} and Figure~\ref{fig:ablation_unpaired_paired} show the comparison of performance between the original unpaired training and our pseudo paired training. As our computation budget reduces, the quality of images generated by the unpaired training method degrades dramatically, while our pseudo paired training method remains relatively stable. The unpaired training requires the model to be strong enough to capture the complicated and ambiguous mapping between the source domain and the target domain. Once the mapping is learned, our student model can learn it from the teacher model directly. Additionally, the student model can still learn extra information on the real target images from the inherited discriminator. 

\renewcommand \arraystretch{1.}
\begin{table}[!t]
\setlength{\tabcolsep}{2.5pt}
\centering
\begin{tabular}{lcccc}
\toprule
Model & ngf & FID & MACs & \#Parameters  \\
\midrule
Original& 64 & 61.75 & 56.8G & 11.38M \\
\midrule
Only change downsample & 64 & 68.72 & 55.5G & 11.13M \\
Only change resBlocks & 64 & 62.95 & 18.3G & 1.98M \\
Only change upsample & 64 & 61.04 & 48.3G & 11.05M \\
\midrule
Only change downsample & 16 & 74.77 & 3.6G & 0.70M \\
Only change resBlocks & 16 & 79.49 & 1.4G & 0.14M \\
Only change upsample & 16 & 95.54 & 3.3G & 0.70M \\
\bottomrule
\end{tabular}
\caption{We report the performance after applying convolution decomposition in each of the three parts (Downsample, ResBlocks, and Upsample) of the ResNet generator respectively on the horse$\to$zebra dataset. \texttt{ngf} denotes the \textbf{n}umber of the \textbf{g}enerator's \textbf{f}ilters. Both the computation and model size are proportional to  \texttt{ngf}$^2$.  We evaluate two settings, \texttt{ngf}=64 and \texttt{ngf}=16. We observe that modifying ResBlock blocks shows a significantly better performance \vs computation trade-off compared to modifying other parts of the network.}
\label{tab:decompose}
\end{table}

\subsubsection{The Effectiveness of Intermediate Distillation and Inheriting the Teacher Discriminator}

\tab{tab:ablation} demonstrates the effectiveness of intermediate distillation and inheriting the teacher discriminator on the Pix2pix model. Solely pruning and distilling intermediate features cannot render a significantly better result than the baseline from-scratch training. We also explore the role of the discriminator in pruning. As a pre-trained discriminator stores useful information of the original generator, it can guide the pruned generator to learn faster and better. If the student discriminator is reset, the knowledge of the pruned student generator will be spoiled by the randomly initialized discriminator, which sometimes yields even worse results than the from-scratch training baseline.

\subsubsection{Effectiveness of Convolution Decomposition}
\label{sec:decomposition}
We systematically analyze the sensitivity of conditional GANs regarding the convolution decomposition transform. We take the ResNet-based generator from CycleGAN to test its effectiveness. We divide the structure of the ResNet generator into three parts according to its network structure: Downsample (3 convolutions), ResBlocks (9 residual blocks), and Upsample (the final two deconvolutions). 
To validate the sensitivity of each stage, we replace all the conventional convolutions in each stage into separable convolutions~\cite{howard2017mobilenets}. The performance drop is reported in Table.~\ref{tab:decompose}. The ResBlock part takes a fair amount of computation cost, so decomposing the convolutions in the ResBlock can notably reduce computation costs. By testing both the architectures with ngf=64 and ngf=16, the ResBlock-modified architecture shows better computation costs \vs performance trade-off. We further explore the computation costs \vs performance trade-off of the ResBlock-modified architecture on the Cityscapes dataset. Figure.~\ref{fig:ablation_original_mobile} illustrates that such MobileNet-style architecture is consistently more efficient than the original one, which has already reduced about half of the computation cost.
\begin{figure}[t]
\centering
\vspace{-5pt}
\includegraphics[width=0.85\linewidth]{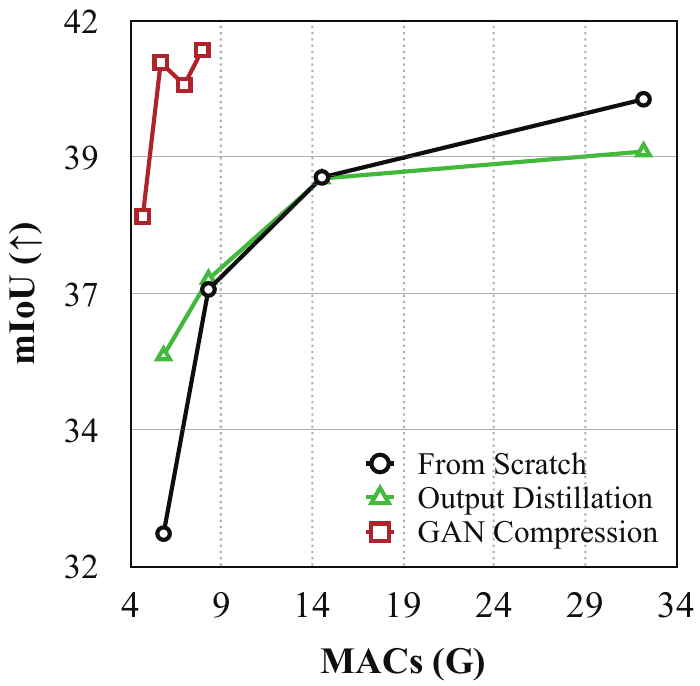}
\vspace{-10pt}
\caption{Performance \vs computation trade-off curve of the Pix2pix model on the Cityscapes dataset~\cite{cordts2016cityscapes}. The output-only  distillation method renders an even worse result than from-scratch training. Our GAN Compression method significantly outperforms these two baselines.}
\label{fig:distillation}
\end{figure}

\renewcommand \arraystretch{1.}
\begin{table}[t]
\setlength{\tabcolsep}{5pt}
\small\centering
\begin{tabular}{lcccc}
\toprule
Method & \multicolumn{2}{c}{MACs} & \multicolumn{2}{c}{mIoU $\uparrow$} \\
\midrule
Original Model & 56.8G & -- & 42.06 & -- \\
From-scratch Training & 5.82G & (9.5$\times$) & 32.57 & (9.49 \frownie{}) \\
\midrule
Aguinaldo \etal~\cite{aguinaldo2019compressing} & 5.82G & (9.5$\times$) & 35.67 & (6.39 \frownie{}) \\
Yim \etal~\cite{yim2017gift} & 5.82G & (9.5$\times$) & 36.69 & (5.37 \frownie{}) \\
Intermediate Distillation & 5.82G & (9.5$\times$) & 38.26 & (\textbf{3.80 \frownie{}}) \\
\midrule
GAN Compression & 5.66G & (\textbf{10.0}$\times$)& 40.77 & (\textbf{1.29 \frownie{}}) \\
\bottomrule
\end{tabular}
\caption{Comparison of GAN Compression and different distillation methods (without NAS) for Pix2pix model on cityscapes dataset. Our intermediate distillation outperforms other methods.}
\vspace{-5pt}
\label{tab:sup_distill}
\end{table}

\subsubsection{Distillation Methods}
\label{sec:distillation}
Recently, Aguinaldo~\etal~\cite{aguinaldo2019compressing} adopted the knowledge distillation to accelerate the unconditional GAN inference. They enforce a student generator's output to approximate a teacher generator's output. However, in the paired conditional GAN training setting, the student generator can already learn enough information from its ground-truth target images. Therefore, the teacher's outputs contain no extra information compared to the ground truth. Figure~\ref{fig:distillation} empirically demonstrates this observation. We run experiments for the Pix2pix model on the Cityscapes dataset~\cite{cordts2016cityscapes}. The results from the distillation baseline~\cite{aguinaldo2019compressing} are even worse than models trained from scratch. Our GAN Compression consistently outperforms these two baselines. We also compare our method with Yim~\etal~\cite{yim2017gift}, a distillation method used in recognition networks. Table~\ref{tab:sup_distill} benchmarks different distillation methods on the Cityscapes dataset for Pix2pix. Our GAN Compression method outperforms other distillation methods by a large margin.
\section{Conclusion}
In this work, we proposed a general-purpose compression framework for reducing the computational cost and model size of generators in conditional GANs. We have used knowledge distillation and neural architecture search to alleviate training instability and to increase the model efficiency. Extensive experiments have shown that our method can compress several conditional GAN models and the visual quality is better preserved compared to competing methods.
\section*{Acknowledgments}

We thank National Science Foundation, MIT-IBM Watson AI Lab, Adobe, Samsung for supporting this research. We thank Ning Xu, Zhuang Liu, Richard Zhang, Taesung Park, and Antonio Torralba for helpful comments. We thank NVIDIA for donating the Jetson AGX Xavier that runs our \href{https://www.youtube.com/playlist?list=PL80kAHvQbh-r5R8UmXhQK1ndqRvPNw_ex}{demo}. 

\ifCLASSOPTIONcaptionsoff
  \newpage
\fi

\bibliographystyle{IEEEtran}
\bibliography{main}

\clearpage
\section{Appendix}
\subsection{Additional Implementation Details}
\subsubsection{Training Epochs}
In all experiments, we adopt the Adam optimizer~\cite{kingma2015adam}. For CycleGAN, Pix2pix and GauGAN, we keep the same learning rate in the beginning and linearly decay the rate to zero over in the later stage of the training. For MUNIT, we decay decays the learning rate by 0.1 every {\ttfamily step\_size} epochs. We use different epochs for the from-scratch training, distillation, and fine-tuning from the once-for-all~\cite{cai2020once} network training. The specific epochs for each task are listed in Table~\ref{tab:hyperparameter}. Please refer to our \href{https://github.com/mit-han-lab/gan-compression}{code} for more details.

\subsubsection{Loss Function}
For the Pix2pix model~\cite{isola2017image}, we replace the vanilla GAN loss~\cite{goodfellow2014generative} by a more stable Hinge GAN loss~\cite{lim2017geometric,miyato2018spectral,zhang2018self}. For the CycleGAN model~\cite{zhu2017unpaired}, MUNIT~\cite{huang2018multimodal} and GauGAN model~\cite{park2019semantic}, we follow the same setting of the original papers and use the LSGAN loss~\cite{mao2017least}, LSGAN loss~\cite{mao2017least} and Hinge GAN loss term, respectively. We use the same GAN loss function for both teacher and student model as well as our baselines. The hyper-parameters $ \lambda_{\text{recon}}$ and $\lambda_{\text{distill}}$ as mentioned in our paper are shown in Table~\ref{tab:hyperparameter}. Please refer to our \href{https://github.com/mit-han-lab/gan-compression}{code} for more details.

\subsubsection{Discriminator}
A discriminator plays a critical role in the GAN training. We adopt the same discriminator architectures as the original work for each model. In our experiments, we did not compress the discriminator as it is not used at inference time. We also experimented with adjusting the capacity of discriminator but found it not helpful. We find that using the high-capacity discriminator with the compressed generator achieves better performance compared to using a compressed discriminator. Table~\ref{tab:hyperparameter} details the capacity of each discriminator.

\subsubsection{Evolutionary Search}
In Fast GAN Compression, we use a much more efficient evolutionary search algorithm. The detailed algorithm is shown in Algorithm~\ref{alg:evolution}.

\renewcommand \arraystretch{1.}
\begin{table*}[t]
\setlength{\tabcolsep}{2.5pt}
\small\centering
\begin{tabular}{cccccccccccccc}
\toprule
\multirow{2}{*}[\multirowcenter]{Model} & \multirow{2}{*}[\multirowcenter]{Dataset} & \multicolumn{2}{c}{Training Epochs} & \multicolumn{2}{c}{Once-for-all Epochs} & \multirow{2}{*}[\multirowcenter]{$\lambda_{\text{recon}}$}& \multirow{2}{*}[\multirowcenter]{$\lambda_{\text{distill}}$} & \multirow{2}{*}[\multirowcenter]{$\lambda_{\text{feat}}$} & \multirow{2}{*}[\multirowcenter]{GAN Loss} &\multicolumn{2}{c}{GAN Compression ngf}& \multirow{2}{*}[\multirowcenter]{ndf} \\
\cmidrule(l{2pt}r{2pt}){3-4} \cmidrule(l{2pt}r{2pt}){5-6} \cmidrule{11-12}
& & Const & Decay & Const & Decay& & & & &Teacher&Student\\
\midrule
CycleGAN & Horse$\to$zebra & 100 & 100 & 200 & 200 & 10  & 0.01& - & LSGAN & 64 & 32 & 64  \\
\midrule
MUNIT & Edges$\to$shoes & 20 & 2 & 40 & 4 & 10  & 1& - & LSGAN & 64 & 48 & 64  \\
\midrule
\multirow{3}{*}{Pix2pix} & Edges$\to$shoes & 5 & 15 & 10 & 30 & 100 & 1   & -  & Hinge & 64 & 48& 128 \\
& Cityscapes & 100 & 150 & 200 & 300 & 100 & 1 & -  & Hinge & 96 & 48 & 128 \\
& Map$\to$arial photo & 100 & 200 & 200 & 400 & 10 & 0.01 & - & Hinge & 96 & 48& 128 \\
\midrule
\multirow{2}{*}{GauGAN} & Cityscapes& 100 & 100 & 100 & 100 & 10  & 10  & 10 & Hinge & 64 & 48 & 64 \\
& COCO-Stuff& 100 & 0 & 100 & 0 & 10  & 10  & 10 & Hinge & -- & -- & 64 \\\bottomrule
\end{tabular}
\caption{Hyper-parameter settings. \textit{Training Epochs} means the epochs for the from-scratch training, distillation and fine-tuning. \textit{Once-for-all Epochs} means epochs for the once-for-all network training. 
\textit{Const} means the epochs of keeping the same initial learning rate (or the total training epochs for MUNIT). \textit{Decay} means epochs of linearly decaying the learning rate to 0 (or the decay {\ttfamily step\_size} for MUNIT).
$\lambda_{\text{recon}}$ and $\lambda_{\text{distill}}$ are the weights of the reconstruction loss term (in GauGAN, this means VGG loss term) and the distillation loss term. $\lambda_{\text{feat}}$ is the weight of the extra GAN feature loss term for GauGAN. \textit{GAN Loss} is the specific type of GAN loss we use for each model. 
\texttt{ngf}, \texttt{ndf} denotes the base number of filters in a generator in and discriminator, respectively, which is an indicator of the model size. Model computation and model size are proportional to \texttt{ngf}$^2$ (or \texttt{ndf}$^2$). For the newest experiment of GauGAN on COCO-Stuff, we directly apply Fast GAN Compression. In Fast GAN Compression, we directly use the original pre-trained model as the teacher, and the {\ttfamily ngf} of the student is the same as the teacher.}

\label{tab:hyperparameter}
\end{table*}
\renewcommand \arraystretch{1.}
\begin{table}[!t]
\setlength{\tabcolsep}{3.5pt}
\small\centering
\begin{tabular}{lccccc}
\toprule
\multirow{2}{*}[\multirowcenter]{Dataset} & \multirow{2}{*}[\multirowcenter]{Arch.} & \multirow{2}{*}[\multirowcenter]{MACs} & \multirow{2}{*}[\multirowcenter]{\#Params} & \multicolumn{2}{c}{Metric}\\
\cmidrule(lr){5-6}
& & & & FID ($\downarrow$) & mIoU ($\uparrow$) \\
\midrule
\multirow{2}{*}{Edges$\to$shoes}& U-net & 18.1G & 54.4M & 59.3 & -- \\
& ResNet & 14.5G & 2.85M & \textbf{30.1} & -- \\
\midrule
\multirow{2}{*}{Cityscapes} & U-net & 18.1G & 54.4M & -- & 28.4 \\
& ResNet & 14.5G & 2.85M & -- & \textbf{33.6}\\
\bottomrule
\end{tabular}
\caption{The comparison of the U-net generator and the ResNet generator for Pix2pix model. The ResNet generator outperforms the U-net generator on both the edges$\to$shoes dataset and Cityscapes dataset.}
\vspace{-5pt}
\label{tab:unet}
\end{table}
\renewcommand \arraystretch{1.}
\begin{table}[!t]
\setlength{\tabcolsep}{4pt}
\small\centering
\begin{tabular}{ccccc}
\toprule
\multirow{2}{*}[\multirowcenter]{Model} & \multirow{2}{*}[\multirowcenter]{Dataset} & \multirow{2}{*}[\multirowcenter]{Setting}  & \multicolumn{2}{c}{Metric} \\
\cmidrule(lr){4-5}
& & & FID ($\downarrow$) & mIoU ($\uparrow$)\\
\midrule
& & Pre-trained & 71.84 & -- \\
CycleGAN & horse$\to$zebra & Retrained & 61.53 & -- \\
& & Compressed & 64.95 & -- \\
\midrule
\multirow{6}{*}[\multirowcenter]{GauGAN}& & Pre-trained & -- & 62.18 \\
 & Cityscapes & Retrained & -- & 61.04 \\
& & Compressed & -- & 61.22 \\
\cmidrule(lr){2-5}
& & Pre-trained & 21.38 & 38.78 \\
 & COCO-Stuff & Retrained & 21.95 & 38.39 \\
& & Compressed & 25.06 & 35.34 \\
\bottomrule
\end{tabular}
\caption{The comparison of the official pre-trained models, our retrained models, and our compressed models. Our retrained CycleGAN model outperforms the official pre-trained models, so we report our retrained model results in Table~\ref{tab:final}. For the GauGAN model, our retrained model with the official codebase is slightly worse than the pre-trained model, so we report the pre-trained model in Table~\ref{tab:final}. However, our compressed model achieves 61.22 mIoU on Cityscapes compared to the pre-trained model.}
\label{tab:retrain}
\vspace{-10pt}
\end{table}

\subsection{Additional Ablation Study}
\subsubsection{Network Architecture for Pix2pix}
For Pix2pix experiments, we replace the original U-net~\cite{ronneberger2015u} by the ResNet-based generator~\cite{johnson2016perceptual}. Table~\ref{tab:unet} verifies our design choice. The ResNet generator achieves better performance on both edges$\to$shoes and cityscapes datasets.

\subsubsection{Retrained Models vs. Pre-trained Models} 
We retrain the original models with minor modifications as mentioned in Section \ref{sec:setups}. Table~\ref{tab:retrain} shows our retrained results. For CyleGAN model, our retrained model slightly outperforms the pre-trained models. For the GauGAN model, our retrained model with official codebase is slightly worse than the the pre-trained model. But our compressed model can also achieve 61.22 mIoU, which has only negligible 0.96 mIoU drop compared to the pre-trained model on Cityscapes.

\begin{algorithm}[b]
\caption{Evolutionary Search}\label{alg:evolution}
\begin{algorithmic}
\Require Population size $P$, mutation ratio $r_m$, parent ratio $r_p$, evaluation iterations $T$, mutation probability $p_m$, computation budget $F$.
\State Mutation size $s_m \gets P \times r_m$
\State Parent size $s_p \gets P \times r_p$
\State $population \gets $ empty array  \Comment The population.
\State $history \gets \varnothing$  \Comment Will contain all sub-networks.
\While{$|population| < P$}  \Comment Initialize population.
    \State $model.arch \gets \Call{RandomArchitecture}{F}$
    \State $model.perf \gets \Call{Eval}{model.arch}$ 
    \State Add $model$ to $population$ and $history$
\EndWhile
\State $i \gets 0$
\While{$i < T$}  \Comment{Evolve for $T$ generations.}
    \State $parents \gets \Call{GetParents}{population, s_p}$ 
    \State $children \gets \text{empty array}$ 
    \State $j \gets 0$
    \While{$j < m_r$} \Comment{Mutate $m_r$ samples}
    \State $arch \gets \Call{Sample}{parents}$ 
    \State $model.arch \gets \Call{Mutate}{arch,p_m,F}$ 
    \State $model.perf \gets \Call{Eval}{model.arch}$ 
    \State Add $model$ to $children$ and $hist$
    \State $j \gets j+1$
    \EndWhile
    \While{$j < P$} \Comment{Crossover $P-m_r$ samples}
    \State $arch_1 \gets \Call{Sample}{parents}$
    \State $arch_2 \gets \Call{Sample}{parents}$
    \State $model.arch \gets \Call{Crossover}{arch_1, arch_2,F}$ 
    \State $model.perf \gets \Call{Eval}{model.arch}$ 
    \State Add $model$ to $children$ and $history$
    \State $j \gets j+1$
    \EndWhile
    \State $population \gets children$
    \State $i \gets i + 1$
\EndWhile
\State \Return best-performed model in $history$
\end{algorithmic}
\end{algorithm}

\subsection{Additional Results}
\begin{figure*}[h]
\centering
\includegraphics[width=\linewidth]{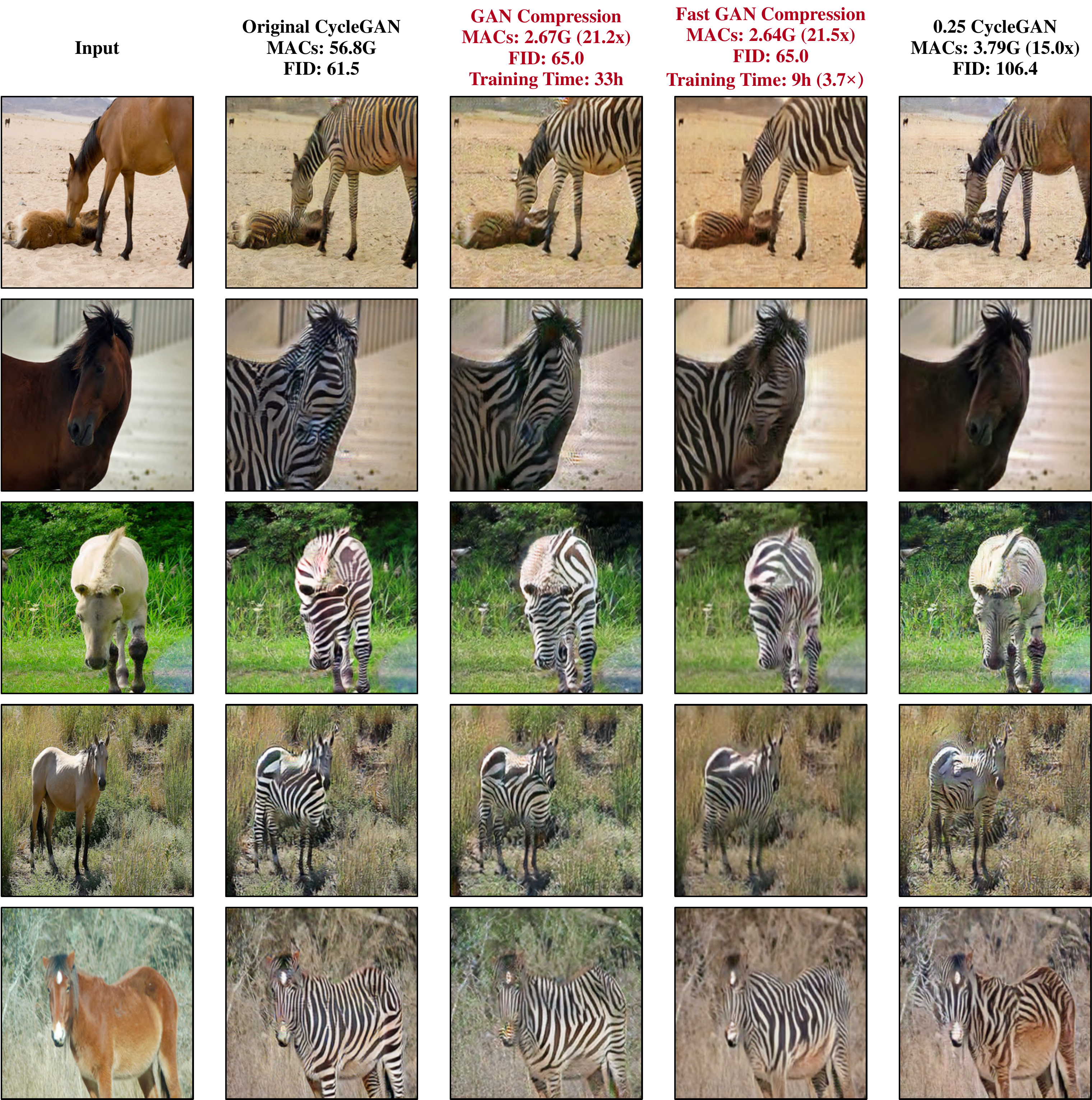}
\caption{Additional results of GAN Compression and Fast GAN Compression with comparison to the 0.25 CycleGAN model on the horse$\to$zebra dataset. Training time is measured on a single 2080Ti GPU. Fast GAN Compression could reduce the training time of GAN Compression by $3.7\times$. }
\label{fig:horse-cyclegan}
\end{figure*}

In Figure~\ref{fig:horse-cyclegan}, we show additional visual results of our proposed GAN Compression and Fast GAN Compression methods for the CycleGAN model in horse$\to$zebra dataset. 

\begin{figure*}[h]
\centering
\includegraphics[width=\linewidth]{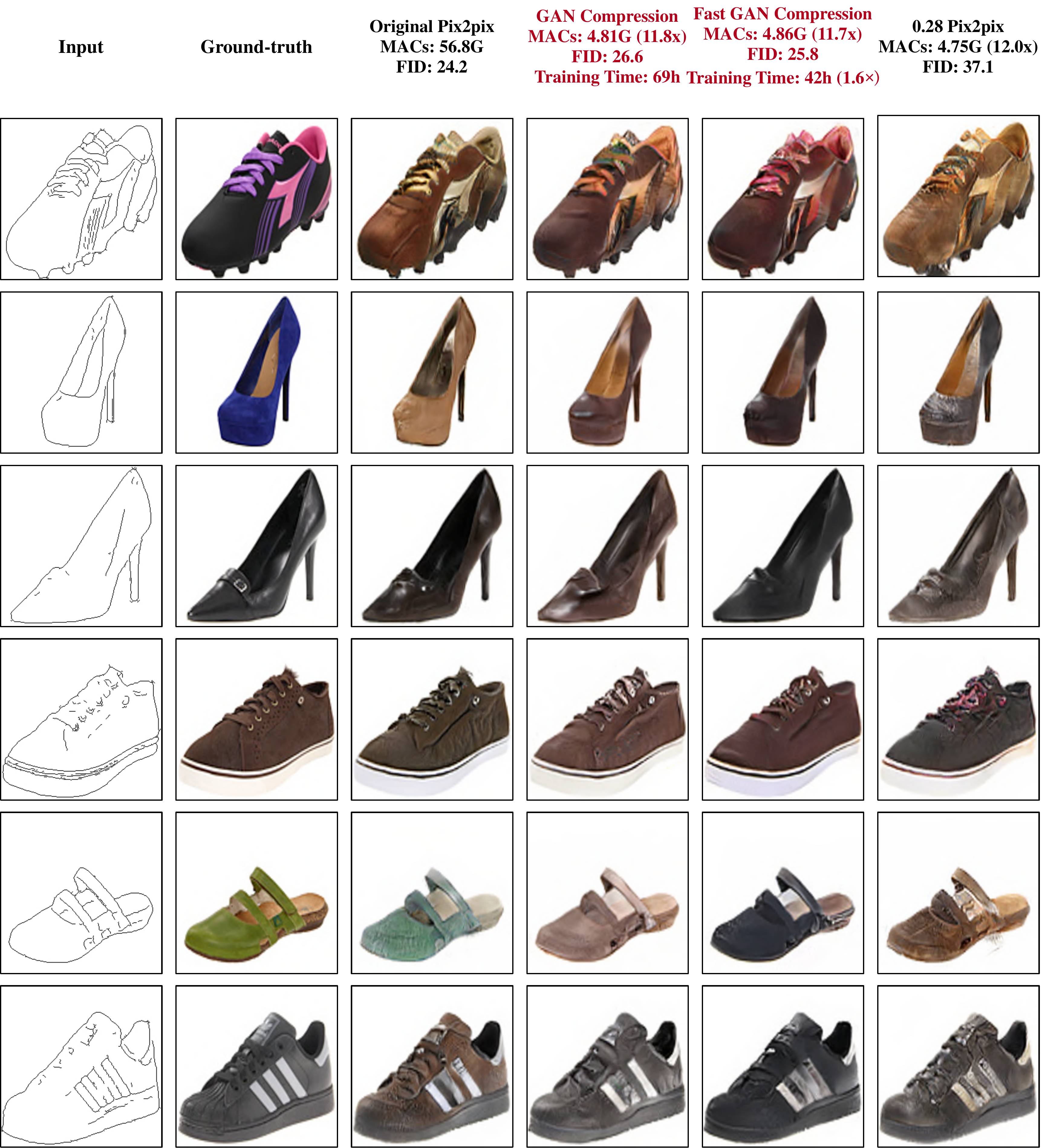}
\caption{Additional results of GAN Compression and Fast GAN Compression with comparison to the 0.28 Pix2pix model on the edges$\to$shoes dataset. Training time is measured on a single 2080Ti GPU. Fast GAN Compression could reduce the training time of GAN Compression by $1.6\times$.}
\label{fig:edges-pix2pix}
\end{figure*}

\begin{figure*}[h]
\centering
\includegraphics[width=\linewidth]{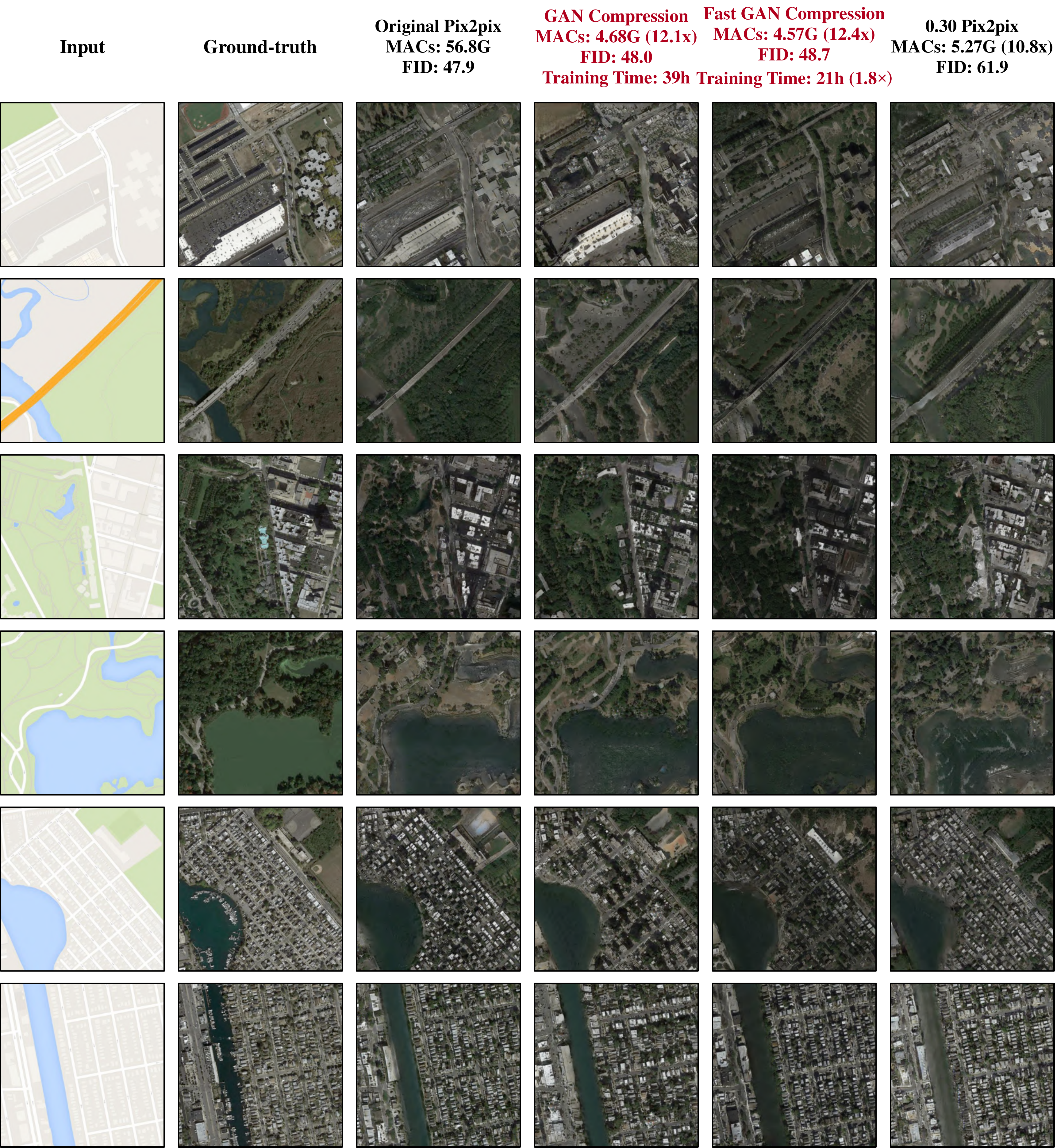}
\caption{Additional results of GAN Compression and Fast GAN Compression with comparison to the 0.30 Pix2pix model on the map$\to$arial photo dataset. Training time is measured on a single 2080Ti GPU. Fast GAN Compression could reduce the training time of GAN Compression by $1.8\times$.}
\label{fig:maps-pix2pix}
\end{figure*}

\begin{figure*}[h]
\centering
\includegraphics[width=\linewidth]{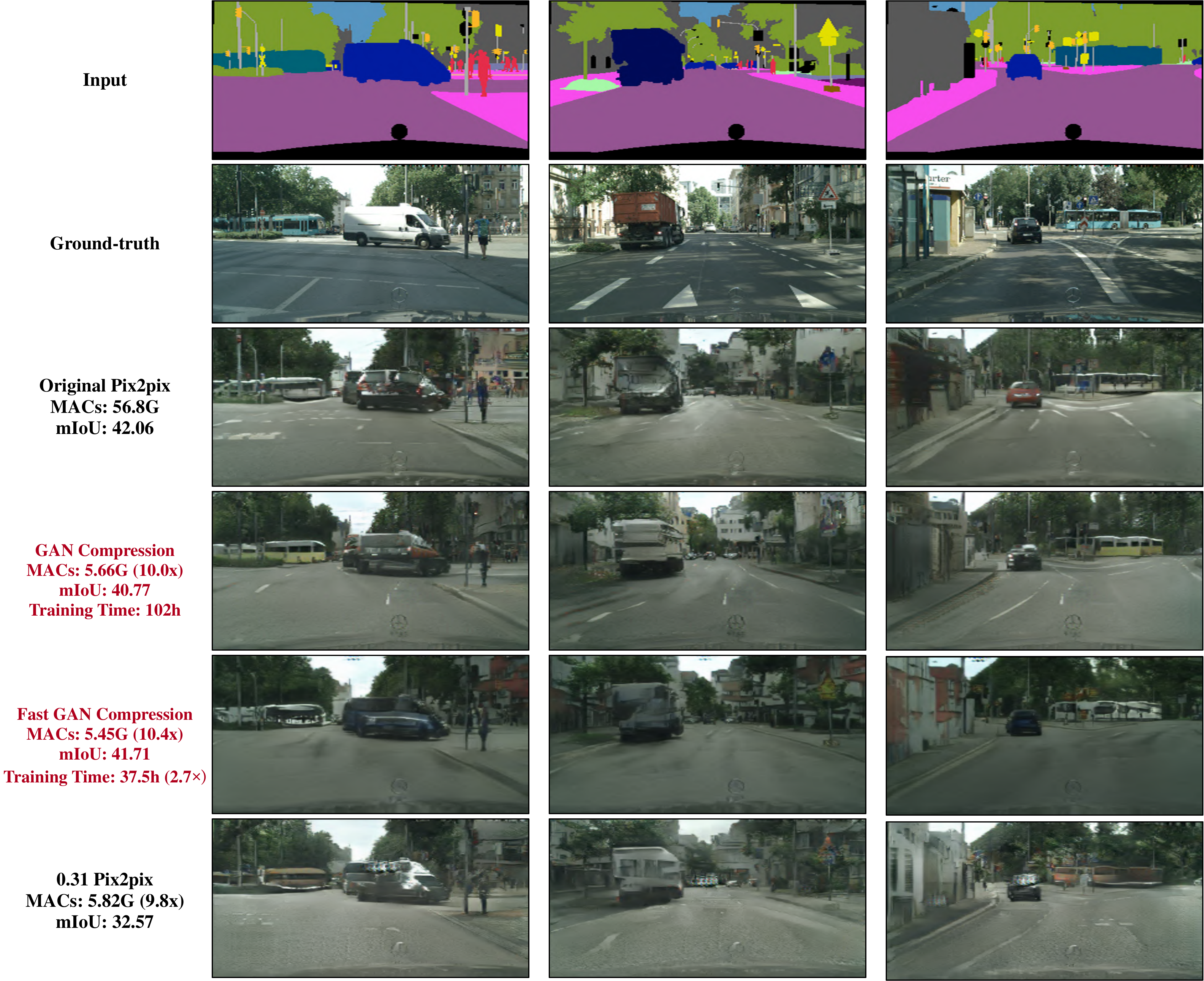}
\caption{Additional results of GAN Compression and Fast GAN Compression with comparison to the 0.31 Pix2pix model on the cityscapes dataset. Training time is measured on a single 2080Ti GPU. Fast GAN Compression could reduce the training time of GAN Compression by $2.7\times$.}
\label{fig:cityscapes-pix2pix}
\end{figure*}

In Figure~\ref{fig:edges-pix2pix}, \ref{fig:maps-pix2pix} and \ref{fig:cityscapes-pix2pix}, we show additional visual results of our proposed methods for the Pix2pix model on edges$\to$shoes, map$\to$arial photo and Cityscapes datasets. 

\begin{figure*}[h]
\centering
\includegraphics[width=\linewidth]{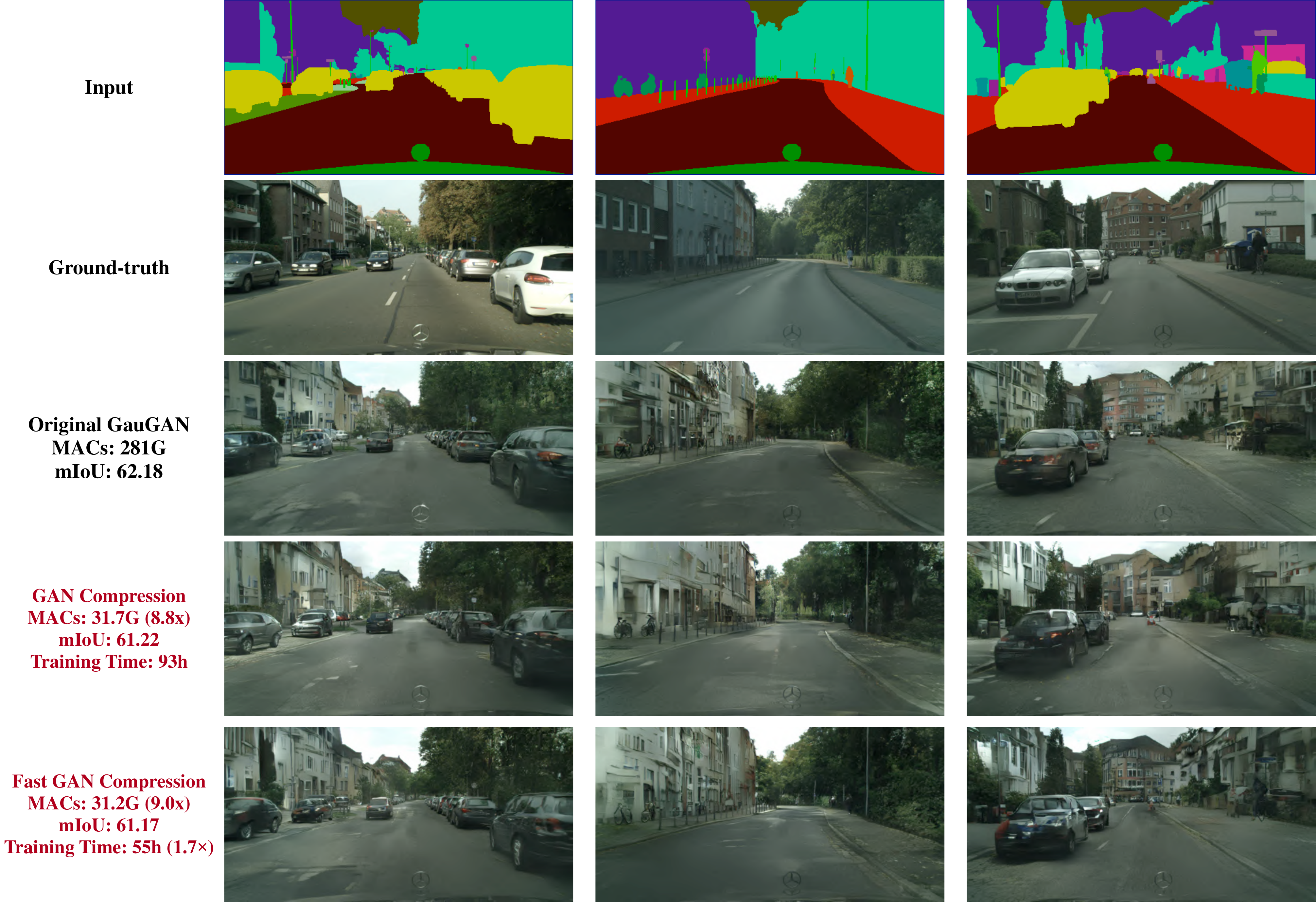}
\caption{Additional results of compressing GauGAN model on the cityscapes dataset. Training time is measured on 8 2080Ti GPUs. Fast GAN Compression could reduce the training time of GAN Compression by $1.7\times$.}
\label{fig:cityscapes-spade}
\end{figure*}

\begin{figure*}[h]
\centering
\includegraphics[width=\linewidth]{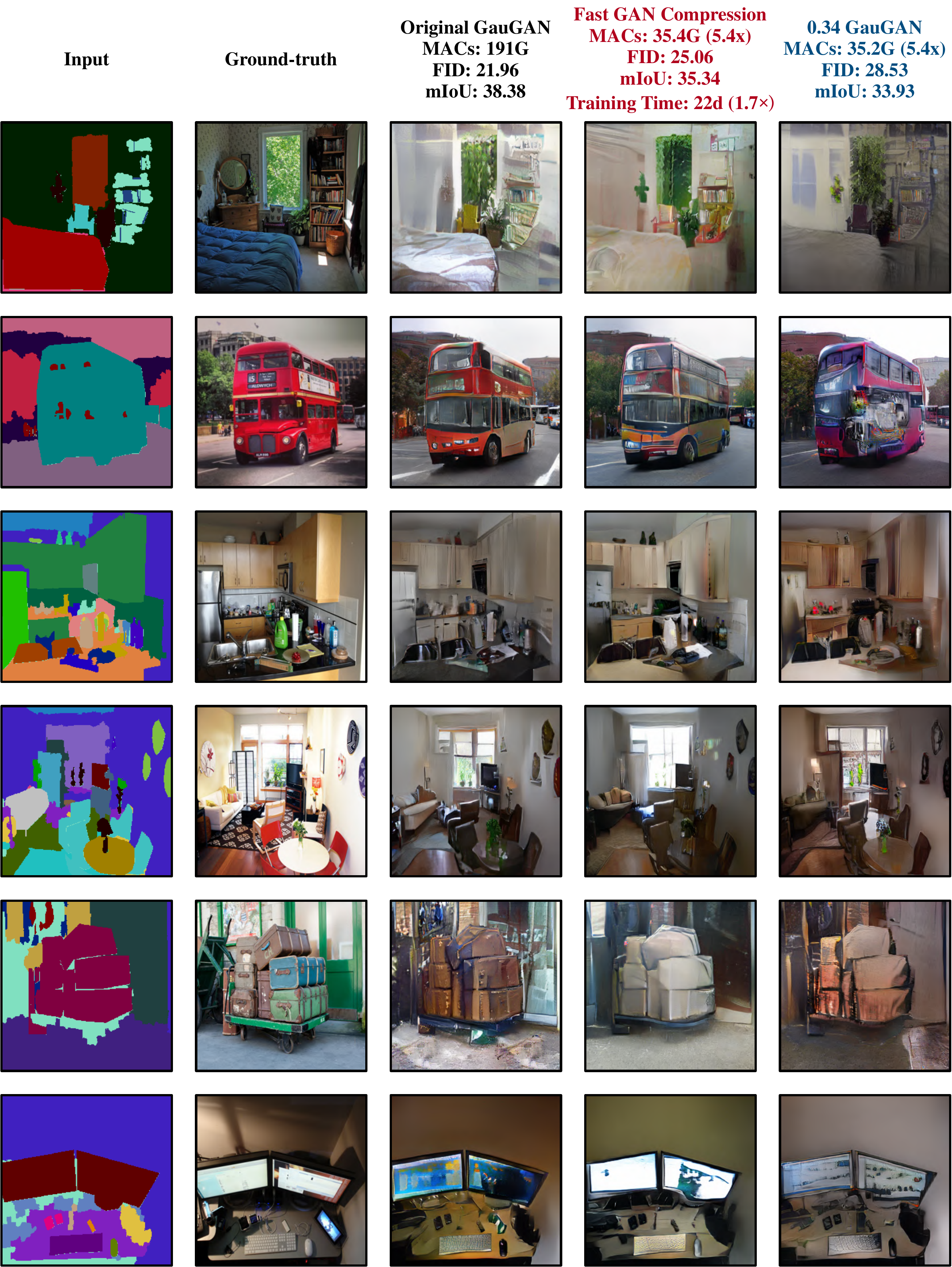}
\caption{Additional results of compressing GauGAN model on COCO-Stuff dataset. Training time is measured on 4 V100 GPUs. It will take about 38 days for GAN Compression on this experiment (as the original model already needs 10 days), which is not affordable, so we directly show the Fast GAN Compression results here. Fast GAN Compression reduces $1.7\times$ training time.}
\label{fig:coco-spade}
\end{figure*}

In Figure~\ref{fig:cityscapes-spade} and \ref{fig:coco-spade}, we show additional visual results of our proposed methods for the GauGAN model on Cityscapes and COCO-Stuff datasets.

\subsection{Changelog}
\noindent \textbf{V1}
 Initial preprint release (CVPR 2020).

\noindent \textbf{V2}
(a) Correct the metric naming (mAP to mIoU). Update the mIoU evaluation protocol of Cityscapes ($\texttt{DRN}(\texttt{upsample}(G(x)))\rightarrow \texttt{upsample}(\texttt{DRN}(G(x)))$). See Section \ref{subsec:metrics} and Table \ref{tab:final} and \ref{tab:ablation}. (b) Add more details regarding Horse2zebra FID evaluation (Section \ref{subsec:metrics}). (c) Compare the official pre-trained models and our retrained models (Table \ref{tab:retrain}).

\noindent \textbf{V3}
(a) Introduce Fast GAN Compression, a more efficient training method with a simplified training pipeline and a faster search strategy (Section~\ref{sec:setups}). (b) Add the results of GauGAN on COCO-Stuff dataset (Table~\ref{tab:final}).

\noindent \textbf{V4}
Accept by T-PAMI. (a) Add experiments on MUNIT (see Table~\ref{tab:final} and Figure~\ref{fig:edges-munit-ref}. (b) Add per-class analysis of GauGAN on Cityscapes.

\end{document}